\documentclass{llncs}

\pdfoutput=1

\usepackage{times}

\usepackage[T1]{fontenc}
\usepackage[algoruled,norelsize,vlined]{algorithm2e}
\usepackage{algorithmic}%
\usepackage[table]{xcolor}
\usepackage{hhline,arydshln}
\usepackage{rotating}
\usepackage{multirow}
\usepackage{subfig}
\usepackage{url}
\usepackage{hyperref}
\usepackage{pict2e}
\usepackage{amsmath, amssymb, amsfonts}
\usepackage{multirow,hhline,arydshln}
\usepackage{color}
\usepackage{pgf}
\usepackage{tikz}
\usepackage{graphicx}
\usepackage{latexsym}
\usepackage{xspace}
\usetikzlibrary{arrows,automata}
\usepackage{epstopdf}
\usepackage{subfig}
\title{\prefixCP Global Constraint for Sequential Pattern Mining}

\newcommand{\topk}[0]{{top-$k$}\xspace}

\def\N{\mbox{$\mathbb{N}$}}

\def\inf{\mbox{$\preceq$}}

\def\prefixCP{\textsc{Prefix-Projection}\xspace}
\def\PP{\texttt{PP}\xspace}
\def\prefixCPREL{\texttt{PP-SRE}\xspace}
\def\prefixCPREG{\texttt{PP-REG}\xspace}

\def\card{\#}

\newcommand\code[1]{{\tt #1}}

\newcommand{\angx}[1]{{{\mbox{$\langle #1 \rangle$}}}}

\newcommand{\SDB}{SDB}	
\newcommand{\I}{\mathcal{I}} 	%ensemble des items
\newcommand{\Li}{\mathcal{L_I}}

\usepackage{etoolbox}
\newbool{seeAll}
\usepackage[normalem]{ulem}

\def\spade{\texttt{SPADE}\xspace}

\def\prefix{\texttt{PrefixSpan}\xspace}
\def\clospan{\texttt{CloSpan}\xspace}

\def\sma{\texttt{SMA}\xspace} 
\def\smap{\texttt{SMA-1P}\xspace} 
\def\smafc{\texttt{SMA-FC}\xspace}

\def\cspade{\texttt{cSpade}\xspace}
\def\gapbide{\texttt{Gap-BIDE}\xspace}
\def\cpsm{\texttt{global-p.f}\xspace}
\def\cps{\texttt{decomposed-p.f}\xspace}

\begin{document}

\title{\prefixCP Global Constraint for Sequential Pattern Mining}
\author{Amina Kemmar$^{1,4}$ \and Samir
  Loudni$^2$ \and Yahia Lebbah$^{1}$\\ Patrice Boizumault$^2$  \and Thierry
  Charnois$^3$} 
\institute{
$^1$LITIO -- University of Oran 1 -- Algeria \\ %- Ahmed Ben Bella, 1524 El-M'Naouer, 31000 Oran, Algeria\\
%$^2$Preparatory school of Oran EPSECG, BP 65 CH 2 Achaba Hnifi, USTO, Oran, Algeria\\
$^2$ GREYC (CNRS UMR 6072) -- University of Caen -- France\\ 
%Campus II C\^ote de Nacre, 14000 Caen - France\\
$^3$LIPN (CNRS UMR 7030) -- University PARIS 13 -- France \\
$^4$Preparatory school of Oran EPSECG -- BP 65 CH 2 Achaba Hnifi, USTO -- Algeria\\
%99, avenue Jean-Baptiste Cl\'ement 93430 Villetaneuse - France
}

\maketitle

%\vspace*{-1.8cm}
\begin{abstract}
Sequential pattern mining under constraints is a
  challenging data mining task. Many efficient ad hoc methods have
  been developed for mining sequential patterns,
but they are all suffering from a lack of genericity.
Recent works have investigated Constraint Programming (CP) methods,
but they are not still effective because of their encoding.
In this paper, we propose a global constraint based on the projected databases principle
which remedies to this drawback.
Experiments show that our approach clearly outperforms CP approaches
and competes well with ad hoc methods on large datasets. 

%%Sequential pattern mining under constraints is a challenging data mining task. 
%%Many efficient ad hoc methods have been developed but they suffer from a lack of genericity.
%%Recent works have investigated Constraint Programming (CP) methods for mining sequential patterns.
%%But,  
%%they are not effective enough because of their CP encoding 
%%and their difficulties to 
%%handle the anti-monotonicity of the frequency measure.
%
%%In this paper, we propose a global constraint based on the projected databases principle
%%which allows to correct these two disadvantages.
%
%%\textcolor{blue}{Experiments show that our approach 
%%clearly outperforms CP approaches
%%and competes well with ad hoc methods on high dimensional datasets for mining frequent sequential patterns 
%%as well as for richer patterns as top-$k$ or regular sequential patterns.
%%}
\end{abstract}
%
%for sequential pattern mining is presented.
%It borrows ideas from the projected databases principle well known in sequential pattern mining. 
% It constrains the values of the next variable to instantiate, to be globally consistent by generating only frequent patterns, with a polynomial time complexity.
% When this constraint is integrated in the machinery of constraint programming, it enables combination with other constraints (e.g., length, regular, top-k), 
% in order to get a declarative framework, performing efficiently rich sequential mining. 
%This contrasts with current sequential pattern mining approaches focused on developing dedicated algorithms for each mining problem. 
%With this declarative approach, we have succeeded to model various sequential mining problems, by combining the global constraint with existing constraints in a CP environment, and then "solve it".
%In our experimental results, while the global constraint alone competes well with classical sequential mining tools on high dimensional datasets, its combination with CP constraints, outperforms classical mining tools on various rich mining problems such as those requiring top-$k$ or regular sequences.

\section{Introduction}
\label{section:introduction}

Mining useful patterns in sequential data is a challenging task.  
Sequential pattern mining is among the most important and popular
data mining task with many real applications such as the analysis of
web click-streams, medical or biological data and textual data.   
For effectiveness and efficiency considerations, many authors have
promoted the use of constraints to focus on the most promising
patterns according to the interests given by the final user.  
In line with \cite{Pei2002}, many efficient ad hoc methods have been 
developed but they suffer from a lack of genericity 
to handle and to push simultaneously sophisticated combination of
various types of constraints. 
Indeed, new constraints have to be hand-coded  and their combinations
often require new implementations.  

Recently, several proposals have investigated relationships between
sequential pattern mining and constraint programming (CP) to revisit
data mining tasks in a declarative and generic 
way~\cite{DBLP:conf/ecai/CoqueryJSS12,metivierLML13,DBLP:conf/ictai/KemmarULCLBC14,NegrevergneCPIAOR15}.   
The great advantage of these approaches is their flexibility. 
The user can model a problem and express his queries by specifying
what constraints need to be satisfied.  
But, all these proposals are not effective enough because of their CP 
encoding. Consequently, the design of new efficient declarative 
models for mining useful patterns in sequential data is 
clearly an important challenge for CP. 

To address this challenge, we investigate in this paper the other side 
of the cross fertilization between data-mining and constraint
programming, namely how the CP framework can benefit from the
power of candidate pruning mechanisms used in sequential pattern mining. 
First,
we introduce the global constraint \prefixCP 
for sequential pattern mining. 
\prefixCP uses a concise encoding
and 
its filtering relies on the principle of projected
databases~\cite{DBLP:conf/icde/PeiHPCDH01}.  
The key idea is to divide the initial database into smaller ones
projected on the frequent subsequences obtained so far,  
then, mine locally frequent patterns in each projected database by
growing a frequent prefix. This global constraint utilizes the
principle of prefix-projected database to keep only locally frequent
items alongside projected databases in order to remove
infrequent ones from the domains of variables.  
Second, we show how the concise encoding allows for a 
straightforward implementation of the frequency constraint (\prefixCP constraint)
and constraints on patterns such as size, item membership 
and regular expressions and the simultaneous combination of them. 
Finally, experiments show that our approach 
clearly outperforms CP approaches
and competes well with ad hoc methods on large datasets for mining
frequent sequential patterns  
or patterns under various constraints.
It is worth noting that the experiments show that our approach
achieves scalability while it is a major issue of CP approaches.  

The paper is organized as follows.
Section~\ref{section:preliminaries} recalls preliminaries.
Section~\ref{PB-related-works} provides a critical review of ad hoc
methods and CP approaches for sequential pattern mining. 
Section~\ref{sec:model} presents the global constraint \prefixCP.
Section~\ref{section:experimentations} reports experiments we
performed. Finally, we conclude and draw some perspectives. 

\section{Preliminaries}
\label{section:preliminaries}
\def\vide{\mbox{$\Box$}} 
\def\ID{\mbox{ID}}

This section presents  background knowledge about sequential pattern
mining  and constraint satisfaction problems. 

\subsection{Sequential Patterns}

Let $\I$ be a finite set of {\em items}. 
The language of sequences  corresponds to $\Li=\I^n$ where $n\in\N^+$. 

\begin{definition}[sequence, sequence database]
A sequence $s$ over $\Li$ is an ordered list
$\langle s_1 s_2 \ldots s_n \rangle$, where $s_i$, $1 \leq i \leq n$,
is an item. $n$ is called the length of the sequence $s$.
A sequence database $\SDB$ is a set of tuples $(sid, s)$,
where $sid$ is a sequence identifier and $s$ a sequence. 
\end{definition}

\begin{definition}[subsequence,  $\inf$ relation]
A sequence $\alpha = \langle \alpha_1 \ldots \alpha_m \rangle$ is a 
  subsequence of $s = \langle s_1 \ldots s_n \rangle$,
denoted by ($\alpha$ $\inf$ $s$), if $m \leq n$ and there exist integers 
$1 \leq j_1 \leq \ldots \leq j_m\leq n$, such that 
$\alpha_i=s_{j_i}$ for all $1 \leq i \leq m$.  We also say that $\alpha$ is
contained in $s$ or $s$ is a super-sequence of $\alpha$. 
For example, the sequence $\angx{BABC}$ is a super-sequence of
$\angx{AC}$ : $\angx{AC}$ $\inf$  $\angx{BABC}$.
A tuple $(sid, s)$  contains a sequence $\alpha$, if $\alpha$ $\inf$ $s$. 
\end{definition}

The cover of a sequence $p$ in $\SDB$ is the set of all tuples in
$\SDB$ in which $p$ is contained. 
The support of a sequence $p$ in $\SDB$  is the number of 
tuples in $\SDB$ which contain $p$. 

\begin{definition}[coverage, support]
Let $\SDB$ be a sequence database and $p$ a sequence. 
$cover_{\SDB}(p)$$=$$\{(sid, s) \in \SDB \, | \, p \,\inf\, s\}$ and
$sup_{\SDB}(p)=\card cover_{\SDB}(p)$. 
\end{definition}

\begin{definition}[sequential pattern]
Given a minimum support threshold $minsup$, every sequence $p$ 
such that $sup_{\SDB}(p) \geq minsup$ is called a sequential pattern~\cite{DBLP:conf/icde/AgrawalS95}. 
$p$ is said to be frequent in $\SDB$. 
\end{definition}

\begin{table}[t]
	\begin{center}
		\scalebox{0.90}{
			\begin{tabular}{|l|l|}	
				\hline
                                ~sid~ &~Sequence~ \\
				\hline
				 $1$ & $\angx{ABCBC}$  \\
				 $2$ & $ \angx{BABC}$ \\
				 $3$ & $ \angx{AB} $ \\
				 $4$  & $\angx{BCD}$ \\
				\hline
			\end{tabular} 
		}  
                        \caption{$\SDB_1$: a sequence database example.}
		\label{tab:SDB}
	\end{center}
\end{table}	

\begin{example}
Table~\ref{tab:SDB} represents
  a sequence database of four sequences 
where the set of items is $ \I = \{A, B, C, D\}$. Let 
the sequence $p= \angx{A C}$. We have
$cover_{{\SDB}1}(p)=\{(1, s_1), (2, s_2)\}$. If we consider 
$minsup=2$, $p=\angx{A C}$ is a sequential pattern because $sup_{\SDB_1}(p) \ge 2$. 
\end{example}

\begin{definition}[sequential pattern mining (SPM)]
Given a sequence database $\SDB$ and a minimum support threshold
$minsup$. The problem of sequential pattern mining is to find all
patterns $p$ such that $sup_{\SDB}(p) \geq minsup$. 
\end{definition}

\subsection{SPM under Constraints}
\label{sec:local}

In this section, we define the problem of mining sequential 
patterns in a sequence database satisfying user-defined
constraints Then, we review the most usual constraints for
the sequential mining problem~\cite{Pei2002}. 

\noindent 
{\bf Problem statement}. Given a constraint $C(p)$ on 
pattern $p$ and a sequence database \SDB, the problem
of constraint-based pattern mining is to find the
complete set of patterns satisfying $C(p)$.
In the following, we present different types of constraints
that we explicit in the context of sequence mining. All
these constraints will be handled by our concise encoding (see
Sections \ref{consistency} and \ref{pattern-encoding}). 

\noindent
- The minimum size constraint $size(p,\ell_{min})$ states that 
the number of items of $p$ must be greater than or equal to $\ell_{min}$. 

\noindent
- The item constraint $item(p,t)$ states that an item $t$ must belong
(or not) to a pattern $p$. 

\noindent
- The regular expression
constraint~\cite{DBLP:journals/tkde/GarofalakisRS02}
$reg(p,exp)$ states that a pattern $p$ must be 
accepted by the deterministic finite 
automata associated to the regular expression $exp$.

\subsection{Projected Databases}
\label{PB-projected-databases}
We now present the necessary definitions related to the concept of
{\it projected databases}~\cite{DBLP:conf/icde/PeiHPCDH01}.

\begin{definition}[prefix, projection, suffix]  
Let $\beta=\angx{\beta_1 \ldots \beta_n}$ and $\alpha = \angx{\alpha_1
  \ldots \alpha_m}$ be two sequences, where $m \leq n$. \\ 
- Sequence $\alpha$ is called the prefix of $\beta$ iff $\forall i \in
[1..m], \alpha_i=\beta_i$.   \\ 
- Sequence $\beta=\angx{\beta_1 \ldots \beta_n}$ is called the 
projection of some sequence $s$ w.r.t. $\alpha$,  
iff (1) $\beta\, \inf\, s$, 
(2) $\alpha$ is a prefix of $\beta$ 
and (3) there exists no proper super-sequence $\beta'$ of $\beta$ such
that $\beta' \,\inf\,s$ and $\beta'$ also has $\alpha$ as prefix. \\
- Sequence $\gamma =\angx{\beta_{m+1} \ldots \beta_n}$ is called the suffix of $s$
w.r.t. $\alpha$. With the standard concatenation operator "$concat$",
we have $\beta = concat(\alpha, \gamma)$.  
\end{definition}

\begin{definition}[projected database] 
Let  $\SDB$ be a sequence database, 
the $\alpha$-projected database, denoted by $SDB|_\alpha$, 
is the collection of suffixes of sequences in  $\SDB$ w.r.t. prefix $\alpha$. 
\end{definition}

\cite{DBLP:conf/icde/PeiHPCDH01} have proposed an
efficient algorithm, called \code{PrefixSpan}, for mining sequential
patterns based on the concept of {\it projected databases}.   
It proceeds by dividing the initial database into smaller ones
projected on the frequent subsequences obtained so far; only their
corresponding suffixes are kept. Then, sequential patterns are mined 
in each projected database by exploring only locally  frequent
patterns. 

\begin{example}
\label{exp-projection}
Let us consider the sequence database of Table~\ref{tab:SDB} with
$minsup=2$.  
\code{PrefixSpan} starts by scanning ${\SDB}_1$ to find
all the frequent items, each of them is used as
a prefix to get projected databases. 
For ${\SDB}_1$, we get $3$ disjoint subsets w.r.t. the prefixes
$\angx{A}$, $\angx{B}$, and $\angx{C}$.  
For instance, ${\SDB}_1|_{\angx{A}}$ consists of 3 suffix sequences:
$\{(1,\angx{B C B C})$, $(2,\angx{BC})$, $(3,\angx{B})\}$.   
Consider the projected database ${\SDB}_1|_{<A>}$, its
locally frequent items are $B$ and $C$. Thus, ${\SDB}_1|_{<A>}$
can be recursively partitioned into 2 subsets w.r.t. the two 
prefixes $\angx{A B}$ and $\angx{A C}$. The $\angx{A B}$- and $\angx{A
  C}$- projected databases can be constructed and recursively mined
similarly. The processing of a 
$\alpha$-projected database terminates  
when no frequent subsequence can be generated. 
\end{example}

Proposition~\ref{prop-prefixspan} establishes the support count of a
sequence $\gamma$ in $\SDB|_\alpha$~\cite{DBLP:conf/icde/PeiHPCDH01}: 
\begin{proposition}[Support count]\label{prop-prefixspan}
For any sequence $\gamma$ in $\SDB$ with prefix $\alpha$ and suffix $\beta$
s.t. $\gamma= concat(\alpha,\beta$), $sup_{\SDB}(\gamma) = sup_{\SDB|_{\alpha}}(\beta)$. 
\end{proposition}

This proposition ensures that only the sequences in $\SDB$ grown from
$\alpha$ need to be considered for the support count of a sequence
$\gamma$. Furthermore, only those suffixes with prefix 
$\alpha$ should be counted.

\subsection{CSP and Global Constraints}

\noindent
A {\it Constraint Satisfaction Problem} (CSP) consists of a 
set $X$ of $n$ variables, a domain $\mathcal{D}$ mapping each variable 
$X_i \in X$ to a finite set of values $D(X_i)$, and a set of constraints
$\mathcal{C}$. An assignment $\sigma$ is a mapping from variables in $X$ to
values in their domains: $\forall X_i \in X, \sigma(X_i) \in D(X_i)$. 
A constraint $c \in \mathcal{C}$ is a subset of the 
cartesian product of the domains of the variables that are in $c$. 
The goal is to find an assignment 
such that all constraints are satisfied.  

\noindent 
\textbf{Domain consistency (DC).} 
Constraint solvers typically use backtracking search to explore the
space of partial assignments. At each assignment, filtering
algorithms prune the search space by enforcing local consistency
properties like domain consistency. A constraint $c$ on $X$ is domain
consistent, if and only if, for every $X_i \in X$ and for every
$d_i \in D(X_i)$, there is an assignment $\sigma$ satisfying $c$ such
that $\sigma(X_i) = d_i$. Such 
an assignment is called a support.  

\noindent 
\textbf{Global constraints} 
provide shorthands to often-used combinatorial substructures.   
We present two global constraints.
Let $X=\angx{X_1, X_2, ..., X_n}$ be a sequence of $n$ variables.

\noindent
Let $V$ be a set of values, $l$ and $u$ be two integers s.t. $0\leq l \leq u \leq n$,
the constraint \texttt{Among}$(X, V ,l, u)$ states that each value $a \in V$ 
should occur at least $l$ times and at most $u$ times in $X$~\cite{DBLP:journals/jmcm/Beldi94}.
Given a deterministic finite automaton $A$,
the constraint \texttt{Regular}$(X, A)$ ensures that the sequence $X$
is accepted by $A$~\cite{regular}.  
%

%\section{Constraint programming}
%\label{sec:cp}
\vspace*{-.15cm}
\section{Related works}
\label{PB-related-works}
This section provides a critical review of ad hoc methods and CP
approaches for SPM. 

\subsection{Ad hoc Methods for SPM}
{\tt GSP}~\cite{DBLP:conf/edbt/SrikantA96} was the first algorithm
proposed to extract sequential patterns. It uses a generate-and test
approach. Later, two major classes of methods have been proposed: 

\noindent -  Depth-first search based on a vertical database format
e.g. \cspade incorporating contraints (max-gap, max-span,
length)~\cite{DBLP:conf/cikm/Zaki00}, 
\spade~\cite{DBLP:journals/ml/Zaki01}   
 or {\tt
   SPAM}~\cite{Ayres:2002:SPM:775047.775109}.

\noindent - Projected pattern growth such as {\tt
  PrefixSpan}~\cite{DBLP:conf/icde/PeiHPCDH01}  
and its extensions, e.g. \clospan\ for mining closed sequential
patterns~\cite{DBLP:conf/sdm/YanHA03} or \gapbide~\cite{Li2012}
tackling the gap constraint. 

In \cite{DBLP:journals/tkde/GarofalakisRS02}, the authors
  proposed {\tt SPIRIT} based on {\tt GSP}
  for SPM with regular expressions. Later, \cite{Bonchi:2008}
  introduces Sequence Mining Automata (\sma), a new approach
  based on a specialized kind of Petri Net. Two variants of \sma
  were proposed:  \smap (\sma one pass) and \smafc (\sma Full
  Check).  \smap processes by means
  of the \sma all sequences one by one, and enters all resulting valid
  patterns in a hash table for support counting, while \smafc allows
  frequency based pruning during the scan of the database. 
Finally, \cite{Pei2002} provides a survey for other constraints 
such as regular expressions, length and aggregates. 
But, 
all these proposals, though efficient, are ad hoc methods suffering
from a lack of genericity. 
Adding new constraints often requires to develop new implementations.

\subsection{CP Methods for SPM}
\label{CP4SPM}

Following the work of \cite{DBLP:journals/ai/GunsNR11} for itemset mining,
several methods have been proposed to mine sequential patterns using CP.

\noindent {\bf Proposals.}
\cite{DBLP:conf/ecai/CoqueryJSS12} have
proposed a first SAT-based model for discovering a special class of
patterns with wildcards\footnote{A wildcard is a special symbol that
  matches any item of $\I$ including itself.} in a single
sequence under different types of 
constraints (e.g. frequency, maximality, closedness). 
{\cite{metivierLML13} have proposed a CSP model for SPM.
Each sequence is encoded by an automaton capturing all subsequences that can occur in it.
\cite{DBLP:conf/ictai/KemmarULCLBC14} have proposed a CSP model for SPM with wildcards.
They show how some constraints dealing with local patterns (e.g. frequency, size, gap, regular expressions)
and constraints defining more complex patterns such as relevant subgroups \cite{DBLP:journals/jmlr/NovakLW09} 
and \topk patterns can be modeled using a CSP. 
\cite{NegrevergneCPIAOR15} have proposed two CP encodings for the SPM. 
The first one uses a global constraint to encode the subsequence relation (denoted \cpsm),
while the second one encodes explicitly this relation using additional
variables and constraints (denoted \cps).

All 
these proposals use {\bf reified constraints} to encode the database.
A reified constraint associates a boolean variable to a constraint
reflecting whether the constraint is satisfied (value 1) or not (value
0).   
For each sequence $s$ of $\SDB$, a reified constraint, 
stating whether (or not) the unknown pattern $p$ is a subsequence of $s$,
is imposed: $(S_s=1) \Leftrightarrow (p \preceq s)$.
A great consequence is that the encoding of the frequency measure 
is straightforward: $freq(p) = \sum_{s \in \SDB} S_s$.
But such an encoding has a major drawback since it
requires $(m=\card\SDB)$ reified constraints to encode the whole
database. This constitutes a strong limitation of the size of the
databases that could be managed. 

Most
of these proposals encode {\bf the subsequence relation} $(p \preceq s)$ 
using variables $Pos_{s,j}$ $(s \in  \SDB$ and $1 \le j \le \ell)$
to determine a position where $p$ occurs in $s$.
Such an encoding requires a large number of additional 
variables ($m$$\times$$\ell$) and makes the labeling computationally
expensive. 
In order to address this drawback, \cite{NegrevergneCPIAOR15} have
proposed a global constraint \texttt{exists-embedding}  
to encode the subsequence relation,
and used projected frequency within an ad hoc specific branching strategy
to keep only frequent  items before branching over the variables of
the pattern. 
But, 
this encoding still relies on reified constraints and requires to
impose $m$ \texttt{exists-embedding} global constraints.

So,
we propose in the next section the \prefixCP global constraint 
that  fully exploits the principle of projected databases to encode
both the subsequence relation and the frequency constraint. 
 \prefixCP does not require any reified constraints nor any extra
 variables to encode the subsequence relation. 
As a consequence, usual SPM constraints (see Section~\ref{sec:local}) 
can be encoded in a straightforward way using directly the (global)
constraints of the CP solver.

\section{\prefixCP Global Constraint}
%\label{section:skypatterncubecomputation}
\label{sec:model}
\newcommand\codex[1]{\mbox{\sc #1}}
This section presents the \prefixCP global
constraint for the SPM problem. 

\subsection{A Concise Encoding}
\label{sec:pattern}

Let $P$ be the unknown pattern of size $\ell$ we are looking for. 
The symbol $\vide$ stands for an empty item and denotes the end of a
sequence.  
The unknown pattern $P$ is encoded with a sequence
of $\ell$ variables $\angx{P_1,P_2,\ldots,P_\ell}$
s.t. $\forall i \in [1\ldots\ell],  D(P_ i)= \I\cup\{\vide\}$.  
There are two basic rules on the domains:
\begin{enumerate}
\vspace*{-.1cm}\item 
To avoid the empty sequence, the first item of $P$ must be non empty,
so $(\vide\not\in D_1)$.  
\item
To allow patterns with less than $\ell$ items, we impose that $\forall
i \in [1.. (\ell$$-$$1)], (P_i=\vide) \rightarrow (P_{i+1} = \vide)$. 
\end{enumerate}

\subsection{Definition and Consistency Checking} 
\label{consistency}

The global constraint \prefixCP ensures both subsequence relation and
minimum frequency constraint.  

\begin{definition}[\prefixCP global constraint]
Let $P = \angx{P_1,P_2,\ldots,P_\ell}$ be a pattern of size $\ell$.   
$\angx{d_1, ..., d_{\ell}} \in D(P_1)\times \ldots \times D(P_\ell)$
is a solution of \prefixCP$(P, \SDB, minsup)$ iff   
$sup_{\SDB}(\angx{d_1, ..., d_{\ell}})
\geq minsup$. 
\end{definition}

\begin{proposition}
\label{prop-solution}
A \prefixCP$(P, \SDB,minsup)$ constraint has a solution if and only if  
there exists an assignment $\sigma = \angx{d_1, ..., d_{\ell}}$ of
variables of $P$ s.t. $\SDB|_{\sigma}$ has at least $minsup$ suffixes of $\sigma$: 
$\card\SDB|_{\sigma}\geq minsup$. 
\end{proposition}

{\it Proof: }
This is a direct consequence of proposition \ref{prop-prefixspan}. We
have straightforwardly   
$sup_{\SDB}(\sigma) = sup_{\SDB|_{\sigma}}(\angx{}) =
\card\SDB|_{\sigma}$. Thus, suffixes of $\SDB|_{\sigma}$ are supports  
of $\sigma$ in the constraint \prefixCP$(P, \SDB,minsup)$,
provided that $\card\SDB|_{\sigma}\geq minsup$. 
$\Box$

The following proposition characterizes values in the domain
of unassigned (i.e. future) variable $P_{i+1}$ that are consistent with the 
current assignment of variables $\angx{P_1, ..., P_i}$.

\begin{proposition}
\label{prop-consistency}
Let $\sigma \footnote{We indifferently denote $\sigma$ by $\angx{d_1,
    \dots, d_i}$ or by $\angx{\sigma(P_1), \dots, \sigma(P_{i})}$.} =\angx{d_1, \dots, d_i}$ be a current
assignment of
variables $\angx{P_1, \dots, P_i}$, $P_{i+1} $ be a future variable. A
value $d \in D(P_{i+1})$ 
appears in a solution for \prefixCP$(P, \SDB, minsup)$ if and only if
$d$ is a frequent item in $\SDB|_{\sigma}$:  
$$\card\{(sid,\gamma) | (sid,\gamma) \in \SDB|_{\sigma} \wedge \angx{d}\inf
\gamma\} \geq minsup$$
\end{proposition}

\noindent
{\it Proof: }
Suppose that value $d \in D(P_{i+1})$ occurs in $\SDB|_{\sigma}$
more than $minsup$. From proposition~\ref{prop-prefixspan}, we have 
$sup_{\SDB}(concat(\sigma, \angx{d})) =
sup_{\SDB|_{\sigma}}(\angx{d})$. Hence, the
assignment $\sigma \cup \angx{d}$ satisfies the constraint,
so $d \in D(P_{i+1})$ participates in a solution. 
$\Box$

%\smallskip
\noindent {\bf Anti-monotonicity of the frequency measure.} 
If a pattern $p$ is not frequent, then any pattern $p'$ satisfying 
$p \,\inf\, p'$ is not frequent.   
From proposition~\ref{prop-consistency} and according 
to the {\it anti-monotonicity property}, we can derive the following 
pruning rule: 

\begin{proposition}
\label{prop-filtering}
Let $\sigma=\angx{d_1, \dots, d_i}$ be a current assignment of
variables $\angx{P_1, \dots, P_i}$. All values $d \in D(P_{i+1})$
that are locally not frequent in
$\SDB|_{\sigma}$ can be pruned from the domain of variable
$P_{i+1}$. Moreover, these values $d$ can also be pruned from the domains
of variables $P_j$ with  $j \in [i+2, \dots, \ell]$. 
\end{proposition}

\noindent
{\it Proof: } 
Let $\sigma=\angx{d_1, \dots, d_i}$ be a current assignment of
variables $\angx{P_1, \dots, P_i}$. Let $d \in D(P_{i+1})$ s.t. 
$\sigma' = {concat(\sigma,\angx{d})}$. 
Suppose that $d$ is not frequent in $\SDB|_{\sigma}$. 
According to proposition \ref{prop-prefixspan},
$sup_{\SDB|_{\sigma}}(\angx{d}) = sup_{\SDB}(\sigma')<minsup$, thus $\sigma'$
is not frequent. So, $d$ can be pruned from the domain of $P_{i+1}$.

\noindent
Suppose that the assignment $\sigma$ has been extended to
$concat(\sigma, \alpha)$, where $\alpha$ corresponds to the assignment
of variables $P_j$ (with $j > i$).  
If $d \in D(P_{i+1})$ is not frequent, it is straightforward that
$sup_{\SDB|_{\sigma}}(concat(\alpha, \angx{d}))\leq
sup_{\SDB|_{\sigma}}(\angx{d}) < minsup$. Thus, if $d$ is not frequent
in $\SDB|_{\sigma}$, it will be also not frequent in
$\SDB|_{concat(\sigma,\alpha)}$. 
So, $d$ can be pruned from the domains of
$P_j$ with $j \in [i+2, \dots, \ell]$.
$\Box$

\begin{example}
Consider the sequence database of Table~\ref{tab:SDB} with
$minsup=2$. Let $P = \angx{P_1,P_2,P_3}$ with $D(P_1) = \I$ and 
$D(P_2) = D(P_3) = \I\cup\{\vide\}$. Suppose that $\sigma(P_1) = A$, 
$\prefixCP(P, \SDB, minsup)$ will remove values $A$ and
$D$ from $D(P_2)$ and $D(P_3)$, since the only locally frequent items in
${\SDB}_1|_{<A>}$ are $B$ and $C$.
\end{example} 

Proposition~\ref{prop-filtering} guarantees 
that any value (i.e. item)  
$d \in D(P_{i+1})$ present but not frequent in $\SDB|_{\sigma}$ does not need
to be considered when extending $\sigma$, thus avoiding searching over
it. 
Clearly, our global constraint encodes the anti-monotonicity
of the frequency measure in a simple and elegant way, while 
CP methods for SPM have difficulties to handle this property. 
In \cite{NegrevergneCPIAOR15}, this is achieved by using very specific
propagators and branching  
strategies, making the integration quite complex (see
\cite{NegrevergneCPIAOR15}).

%%%%%%%%%%%%%%%%%%%%%%%%%%%%%%%%%%%%%%%%%%%%%%%%%%%%%%%%%%%%%%%%%%%%%%%%%%%%%%%%%%%
\begin{algorithm}[t]
\begin{small}
\caption{\small \codex{ProjectSDB}($\SDB$, $ProjSDB$, $\alpha$)  \label{algo:projection}}
\KwData{$\SDB$: initial database; $ProjSDB$: projected sequences; $\alpha$: prefix}
\Begin
{
\lnl{p1}$\SDB|_{\alpha} \leftarrow \emptyset$ \;
\lnl{p2}\For{each pair $(sid,start) \in ProjSDB$}{
\lnl{p3}$s \leftarrow \SDB[sid]$ \; 
\lnl{p4}$pos_{\alpha}  \leftarrow 1$;\ $pos_s  \leftarrow start$ \;
\lnl{p6}\While{$(pos_{\alpha} \leq \card\alpha$ $\wedge$ $pos_s \leq \card s)$}{
\lnl{p7}\If{$(\alpha[pos_\alpha] = s[pos_s])$}{
\lnl{p8}$pos_\alpha \leftarrow pos_\alpha + 1$ \;
}
\lnl{p9}$pos_s \leftarrow pos_s + 1$ \;

}
\lnl{p10}\If{$(pos_\alpha = \card\alpha + 1)$}{
\lnl{p11}$\SDB|_{\alpha} \leftarrow \SDB|_{\alpha} \cup \{(sid, pos_s)\}$
}
}
\lnl{p12}\Return $\SDB|_{\alpha}$ \;
}
\end{small}
\end{algorithm}
%%%%%%%%%%%%%%%%%%%%%%%%%%%%%%%%%%%%%%%%%%%%%%%%%%%%%%%%%%%%%%%%%%%%%%%%%%%%%%%%%%%

\subsection{Building the projected databases.} 
\label{sec-ProjectSDB}
The key issue of our approach lies in the construction of the 
projected databases. 
When projecting a prefix, instead of storing the whole suffix as a
projected subsequence, one 
can represent each suffix by a pair $(sid, start)$ where $sid$ is the
sequence identifier and $start$ is the starting position of the
projected suffix in the 
sequence $sid$. For instance, let us consider the sequence database of Table~\ref{tab:SDB}. As
shown in example \ref{exp-projection}, ${\SDB}|_{\angx{A}}$
consists of 3 suffix sequences: $\{(1,\angx{B C 
  B C})$, $(2,\angx{BC})$, $(3,\angx{B})\}$. 
By using the {\it pseudo-projection}, ${\SDB}|_{\angx{A}}$ can be
represented by the following three pairs: $\{(1,2)$, $(2,3)$, $(3,2)\}$. 
This is the principle of {\it pseudo-projection}, adopted in \prefix, 
exploited during the filtering step of our \prefixCP global
constraint. Algorithm \ref{algo:projection} details 
this principle. It takes as input a set of projected sequences
$ProjSDB$ and a prefix $\alpha$.  
The algorithm processes all the pairs $(sid,start)$ of $ProjSDB$ one
by one (line \ref{p2}), and searches for the lowest location of
$\alpha$ in the sequence $s$ corresponding to the $sid$ of that
sequence in $\SDB$ (lines \ref{p7}-\ref{p9}). 

In the worst case, \codex{ProjectSDB} processes all the items of all
sequences. So, the time complexity is $O(\ell\times m)$, with  
$m=\card\SDB$ and $\ell$ is the length of the longest sequence in 
$\SDB$. 
The worst case space complexity of pseudo-projection is $O(m)$, since
we need to store for each sequence only a pair ($sid, start$), while
for the standard projection the space complexity is $O(m\times
\ell)$. Clearly, the pseudo-projection takes much less space than the standard projection.

\begin{algorithm}[!ht]
\begin{small}
\SetKwFunction{ProcProjection}{\codex{Function ProjectSDB}}
\SetKwFunction{ProcGetFreqItems}{\codex{Function getFreqItems}}
\caption{\small \codex{Filter-Prefix-Projection}($\SDB$, $\sigma$, $i$, $P$, $minsup$) \label{algo:filter}}
\KwData{$\SDB$: initial database; $\sigma$: current prefix
  $\angx{\sigma(P_1), \ldots,\sigma(P_i)}$; $minsup$: the minimum
  support threshold; $\mathcal{PSDB}$: internal data structure of
  \prefixCP for storing pseudo-projected databases} 
\Begin
{
\lnl{pre1}\If{$( i \geq 2 \wedge \sigma(P_i) = \vide)$}{ 
                  \lnl{pre2}\For{$j \leftarrow i+1$ \KwTo $\ell$}{
                                  \lnl{pre3}$P_{j}  \leftarrow \vide$\;
                                 }
                   \lnl{pre4} \Return True\;
                   }
\Else{
          \lnl{pre0} $\mathcal{PSDB}_ {i} \leftarrow
          \codex{ProjectSDB}(\SDB, \mathcal{PSDB}_{i-1},
          \angx{\sigma(P_{i})})$\;
          \lnl{proj5}\If{$(\card\mathcal{PSDB}_{i} < minsup)$}{
            \lnl{proj6} \Return False \;
          }
          \Else{
                    \lnl{proj7} $\mathcal{FI} \leftarrow
                    \codex{getFreqItems}(\SDB, \mathcal{PSDB}_{i}, minsup)$ \;
                     \lnl{proj8}\For{$j \leftarrow i+1$ \KwTo $\ell$}{
                                 \lnl{proj9} \ForEach{$a \in D(P_j) \, s.t. (a \neq \vide \wedge a \notin \mathcal{FI} )$}{
                                   \lnl{proj10}$D(P_{j}) \leftarrow
                                   D(P_{j}) - \{a\}$\;
                                 }
                                  }
                       \lnl{proj13} \Return True\;
          }
     }
}

\ProcGetFreqItems($\SDB$, $ProjSDB$, $minsup$) \;
 \KwData{$SDB$: the initial database; $ProjSDB$: pseudo-projected
   database; $minsup$: the minimum support threshold; 
   $ExistsItem$, $SupCount$: internal data structures using a hash table
   for support counting over items;}
\Begin
{
\lnl{F1} $SupCount[ ] \leftarrow \{0, ..., 0\}$;\ \ $F\leftarrow \emptyset$ \;
\lnl{F2}\For{each pair $(sid,start) \in ProjSDB$}{
\lnl{F3} $ExistsItem[] \leftarrow \{false, ..., false\}$; $s \leftarrow SDB[sid]$ \;
\lnl{F4}\For{$i \leftarrow start$ \KwTo $\card s$}{
\lnl{F5}$a \leftarrow s[i]$ \;

\lnl{F6}\If{$(\neg ExistsItem[a])$}{
\lnl{F7}$SupCount[a] \leftarrow SupCount[a] + 1$ \;
 \lnl{F8}$ExistsItem[a] \leftarrow true$\;
\lnl{F10}\If{$(SupCount[a] \geq minsup)$}{
\lnl{F11}$F \leftarrow F \cup \{a\}$\;
}
}
}
}
\lnl{F12}\Return $F$\;
}
\end{small}
\end{algorithm}
%%%%%%%%%%%%%%%%%%%%%%%%%%%%%%%%%%%%%%%%%%%%%%%%%%%%%%%%%%%%%%%%%%%%%%%%%%%%%%%%%%%

\subsection{Filtering}
\label{Filtering}

Ensuring DC on $\prefixCP(P, \SDB, minsup)$ is equivalent to finding a sequential pattern of length $(\ell-1)$ and 
then checking whether this pattern remains a frequent pattern 
when extended to any item $d_{\ell}$ in $D(P_{\ell})$. 
Thus, finding such an assignment (i.e. support) is as much as difficult than the original 
problem of sequential pattern mining. 
\cite{Guizhen-2006} has proved that the problem of 
counting the number of maximal\footnote{A sequential pattern $p$ is
maximal if there is no sequential pattern $q$ such that $p \inf q$.}
frequent patterns in a database of sequences is \#P-complete, thereby
proving the NP-hardness of 
the problem of mining maximal frequent sequences. 
The difficulty is due to the exponential
number of candidates that should 
be parsed to find the frequent patterns. Thus, finding, for every
variable $P_i \in
P$ and for every $d_i \in D(P_i)$, an assignment $\sigma$ satisfying
$\prefixCP(P, \SDB, minsup)$ s.t. $\sigma(P_i) = d_i$ is of 
exponential nature.  

So, the filtering of the \prefixCP constraint maintains a consistency lower than DC.
This consistency is based on specific properties of the projected
databases (see Proposition~\ref{prop-consistency}), and
anti-monotonicity of the frequency constraint (see
Proposition~\ref{prop-filtering}), 
and resembles forward-checking regarding Proposition~\ref{prop-consistency}. 
\prefixCP is considered as a global constraint, since all variables
share the same internal data structures that awake and drive the
filtering. 

Algorithm~\ref{algo:filter} describes the  pseudo-code of the
filtering algorithm of the \prefixCP constraint.  
It is an incremental filtering algorithm
that should be run when some $i$ first variables are
assigned according to the following lexicographic ordering $\angx{P_1,
  P_2, \dots, P_\ell}$ of variables of $P$. 
It exploits internal data-structures enabling to enhance the
filtering algorithm. 
More precisely, it uses an incremental data structure, denoted
$\mathcal{PSDB}$, that stores the intermediate pseudo-projections of
$\SDB$, where $\mathcal{PSDB}_i$ ($i\in [0, \ldots, \ell]$)
corresponds to the $\sigma$-projected 
database of the current partial assignment $\sigma=\angx{\sigma(P_1),
  \ldots, \sigma(P_i)}$ (also called prefix) of variables $\angx{P_1,
  \dots, P_i}$, and $\mathcal{PSDB}_0 = \{(sid, 1)| (sid, s) \in
\SDB\}$ is the initial pseudo-projected database of $\SDB$ 
(case where $\sigma = \angx{}$). 
It also uses a hash table indexing the items $\I$ into integers
$(1 \ldots \card\I)$ for an efficient support counting over items
(see function {\tt getFreqItems}).  

Algorithm~\ref{algo:filter} takes as input the current partial
assignment $\sigma=\angx{\sigma(P_1), \ldots, \sigma(P_i)}$
of variables $\angx{P_1, \dots, P_i}$, 
the length $i$ of $\sigma$ (i.e. position of the last assigned
variable in $P$) and the minimum support threshold
$minsup$. 
It starts by checking if the last assigned variable
$P_i$ is instantiated to $\vide$ (line \ref{pre1}). In this case, the end
of sequence is reached (since value $\vide$ can only appear at the
end) and the sequence $\angx{\sigma(P_1), \dots, \sigma(P_{i})}$
constitutes a frequent pattern in $\SDB$; hence the algorithm sets the 
remaining $(\ell - i)$ unassigned variables to $\vide$ and returns
{\it true}  (lines \ref{pre2}-\ref{pre4}). Otherwise, the algorithm
computes incrementally $\mathcal{PSDB}_i$ from $\mathcal{PSDB}_{i-1}$ 
by calling function \codex{ProjectSDB} (see
Algorithm~\ref{algo:projection}).  Then, it checks in line
\ref{proj5} whether the current assignment $\sigma$ is a {\it legal} prefix
for the constraint (see Proposition \ref{prop-solution}). This is
done by computing the size of $\mathcal{PSDB}_i$. If this size is less
than $minsup$, we stop growing $\sigma$ and we return {\it
  false}. 
Otherwise, the algorithm computes the set of
  locally frequent items $\mathcal{F_I}$ in $\mathcal{PSDB}_i$
by calling function {\tt getFreqItems} (line
\ref{proj7}). 

Function  {\tt getFreqItems} processes all the entries of the
pseudo-projected database 
one by one, counts the number of first occurrences of items
$a$ (i.e. $SupCount[a]$) in each entry $(sid, start)$, and keeps only
the frequent ones (lines \ref{F1}-\ref{F11}). 
This is done by using $ExistsItem$ data structure. 
After the whole pseudo-projected database has been processed, the
frequent items are returned (line \ref{F12}), and Algorithm~\ref{algo:filter} 
updates the current domains of variables $P_j$ with $j \geq 
(i+1)$ by pruning inconsistent values, 
thus avoiding searching over not frequent items (lines
\ref{proj8}-\ref{proj10}). 

\begin{proposition}\label{prop-complexity}
In the worst case, filtering with \prefixCP global constraint can be
achieved in $O(m\times\ell + m\times d + \ell\times d)$. The worst
case space complexity of \prefixCP is $O(m\times \ell)$. 
\end{proposition}

\noindent
{\it Proof: }
Let $\ell$ be the length of the longest sequence in $\SDB$, $m$ $=$
$\card\SDB$, and $d$ $=$ $\card\I$. 
Computing the pseudo-projected database $\mathcal{PSDB}_i$ can be
done in $O(m \times \ell)$: for 
each sequence $(sid,s)$ of $\SDB$, checking if $\sigma$ occurs in $s$
is $O(\ell)$ and there are $m$ sequences. The total complexity of
function \codex{GetFreqItems} is $O(m\times(\ell + d))$.  
Lines (\ref{proj8}-\ref{proj10}) can be achieved in $O(\ell\times d)$.
So, the whole complexity is $O(m \times \ell + m\times(\ell + d) +
\ell\times d)$ $=$ $O(m\times\ell + m\times d + \ell\times d)$. 
The space complexity of the filtering algorithm lies in the storage of
the $\mathcal{PSDB}$ internal data structure. In the worst case, we have to
store $\ell$ pseudo-projected databases. Since each pseudo-projected database
requires $O(m)$, the worst case space complexity is $O(m\times \ell)$.  
$\Box$

\subsection{Encoding of SPM Constraints}
\label{pattern-encoding}
Usual SPM constraints (see Section~\ref{sec:local}) can be reformulated in a straightforward way.
Let $P$ be the unknown pattern.
 
\noindent- {\it Minimum size constraint}:
$size(P,\ell_{min})  \equiv
\bigwedge_{i=1}^{i=\ell_{min}} (P_i \neq \square)$

\noindent- {\it Item constraint}: let $V$ be a subset of items, $l$ and $u$
two integers  s.t. $0 \leq l\leq u \leq \ell$.  
\hspace*{-.07cm}$item(P,V) \equiv \bigwedge_{t \in V} \mbox{\tt Among}(P,\{t\},l,u)$
enforces that items of $V$ should occur at least $l$ times and at most $u$ times in $P$.
To forbid items of $V$ to occur in $P$, $l$ and $u$ must be set to $0$. 

\noindent 
- {\it Regular expression constraint}:
let $A_{reg}$ be the deterministic finite automaton encoding the regular
expression $exp$.
$reg(P,exp) \equiv {\tt Regular}(P, A_{reg})$.

%\section{Mining the \topk sequential patterns}
%\label{section:useful:patterns}
%\input{tex/5-topk.tex}

\section{Experimental Evaluation}
\label{section:experimentations}
This section reports experiments on several real-life
datasets from~\cite{JMLR:v15:fournierviger14a,BCCC2012cbms,Bonchi:2008} 
of large size having varied characteristics and representing
different application domains 
(see Table~\ref{table:CharData}). 
Our objective is
(1) to compare our approach to
existing CP methods as well as to state-of-the-art methods for SPM 
in terms of scalability which is a major issue of existing CP
methods, 
(2) to show the flexibility of our approach allowing to handle different
constraints simultaneously. 

\begin{table}[t] \centering
\scalebox{0.85}{
\begin{tabular}{|l|c|c|c|c|c|}
\hline
dataset & $\card\SDB$ & $\card\I$ & avg $(\card s)$ & $\max_{s\in
                                                      \SDB}$ $(\card
                                                      s)$ & type of
                                                            data\\ 
\hline
%\hline
%Snake & 163 & 20 & 60 & protein sequences\\
%\hline
%Sign & 730 & 267 & 51.99 & 0.19474 & sign language utterances\\
%\hline
Leviathen & 5834 & 9025 & 33.81 & 100 & book\\
%\hline
\hline
Kosarak & 69999 & 21144 & 7.97 & 796   & web click stream\\ 
\hline
FIFA & 20450 & 2990 & 34.74 & 100 & web click stream \\
\hline
BIBLE & 36369 & 13905 & 21.64 & 100 & bible \\
%\hline
%BMS & 59,601 & 497 & 2.42  & 0.0050& web click stream \\
%\hline
%MSNBC & 31,790 & 17 & 13.33  & 0.78414& web click stream \\
\hline
Protein & 103120 &  24 &  482 & 600 & protein sequences \\
\hline
data-200K & 200000 & 20 & 50 & 86 & synthetic dataset\\
\hline
PubMed & 17527 & 19931 &  29 & 198 & bio-medical text\\
\hline
\end{tabular}
}
%\vspace{-0.35cm}
\caption{Dataset Characteristics.} 
\label{table:CharData}
\end{table}

\subsection{Experimental protocol} 
The implementation of our approach was carried out in the {\tt Gecode}
solver\footnote{\url{http://www.gecode.org}}. 
All experiments were conducted on a machine with a processor Intel X5670 and 24 GB of memory. 
%running the Linux 
%operating system, {\color{blue}except for SPM under regular
%expressions when we have used windows system (Intel i5-3210M of 2.50 GHz with 8GB memory) in order to compare with \sma for which 
%the a%uthors provide only the windows executable file}.  
A time limit of 1 hour has been used. 
For each dataset, we varied the $minsup$ threshold until the
methods are not able to complete the extraction of all patterns
within the time limit. 
$\ell$ was set to the length of the longest
sequence of $\SDB$. The implementation and the datasets used in our
experiments are available
online\footnote{\url{https://sites.google.com/site/prefixprojection4cp/}}. 
We compare our approach (indicated by \PP) with:
%\vspace*{-.8cm}
\begin{enumerate} 
\item two CP encodings~\cite{NegrevergneCPIAOR15}, the most efficient CP
  methods for SPM: \cpsm and 
  \cps; 
\item state-of-the-art methods for SPM : \prefix and \cspade; 
\item \sma~\cite{Bonchi:2008} for SPM under regular expressions. 
\end{enumerate}

We used the author's \cspade
implementation~\footnote{\url{http://www.cs.rpi.edu/~zaki/www-new/pmwiki.php/Software/}} 
for SPM, the publicly available implementations of \prefix by
Y. Tabei~\footnote{\url{https://code.google.com/p/prefixspan/}} and 
the \sma
implementation~\footnote{\url{http://www-kdd.isti.cnr.it/SMA/}} for
SPM under regular expressions. The
implementation~\footnote{\url{https://dtai.cs.kuleuven.be/CP4IM/cpsm/}}
of the two CP encodings was carried out in the {\tt Gecode} solver. 
All methods have been executed on the same machine.

%%%%%%%%%%%%%%%%%%%%%%%%%%%%%%%%%%%%%%%%%%%%%%%%%%%%%%%%%%%%%%%%%
\begin{figure*}[t]	
{\footnotesize
\begin{tabular}{ccc}
BIBLE & Kosarak & PubMed \\
	\includegraphics[width=3.8cm, height=2.8cm]{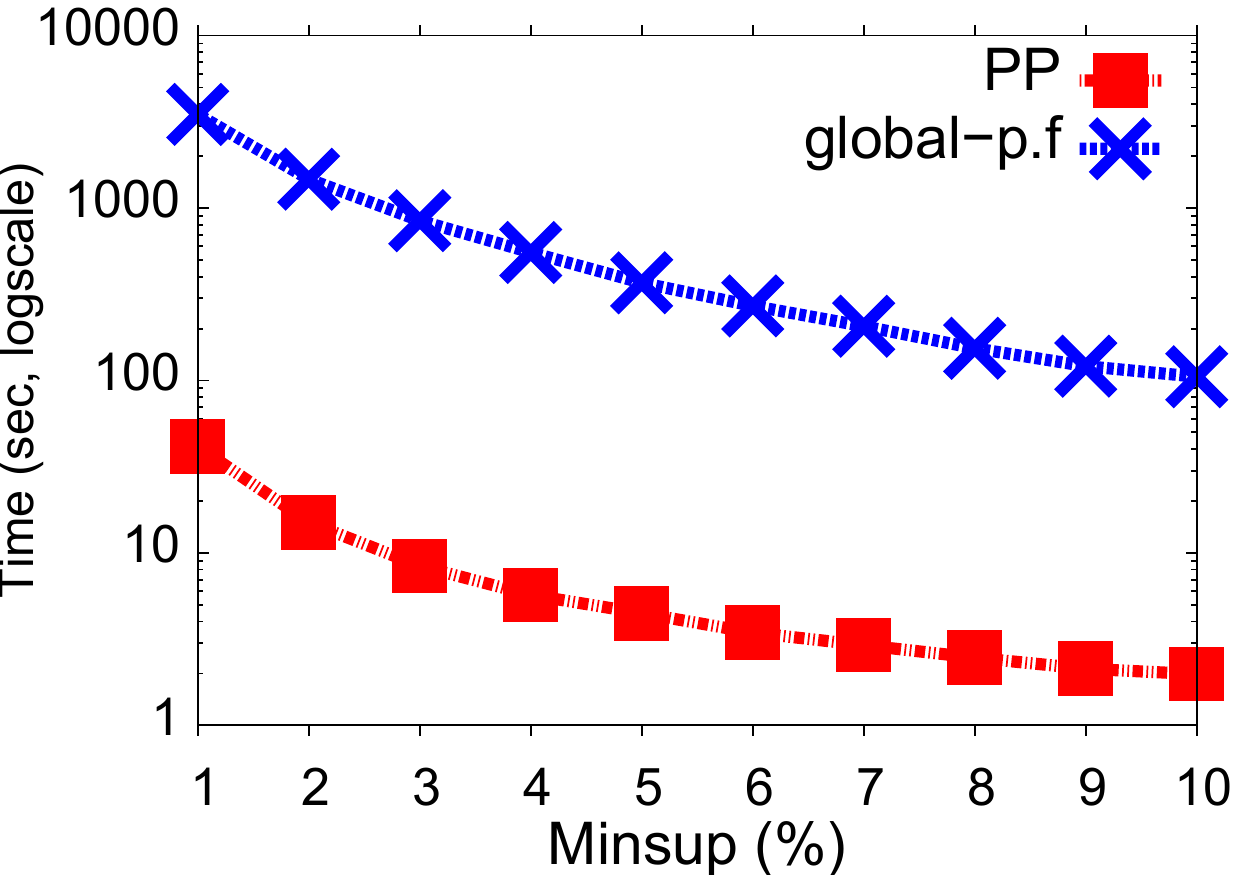} %Freq-BIBLE-col.pdf}     
&   
\includegraphics[width=3.8cm, height=2.8cm]{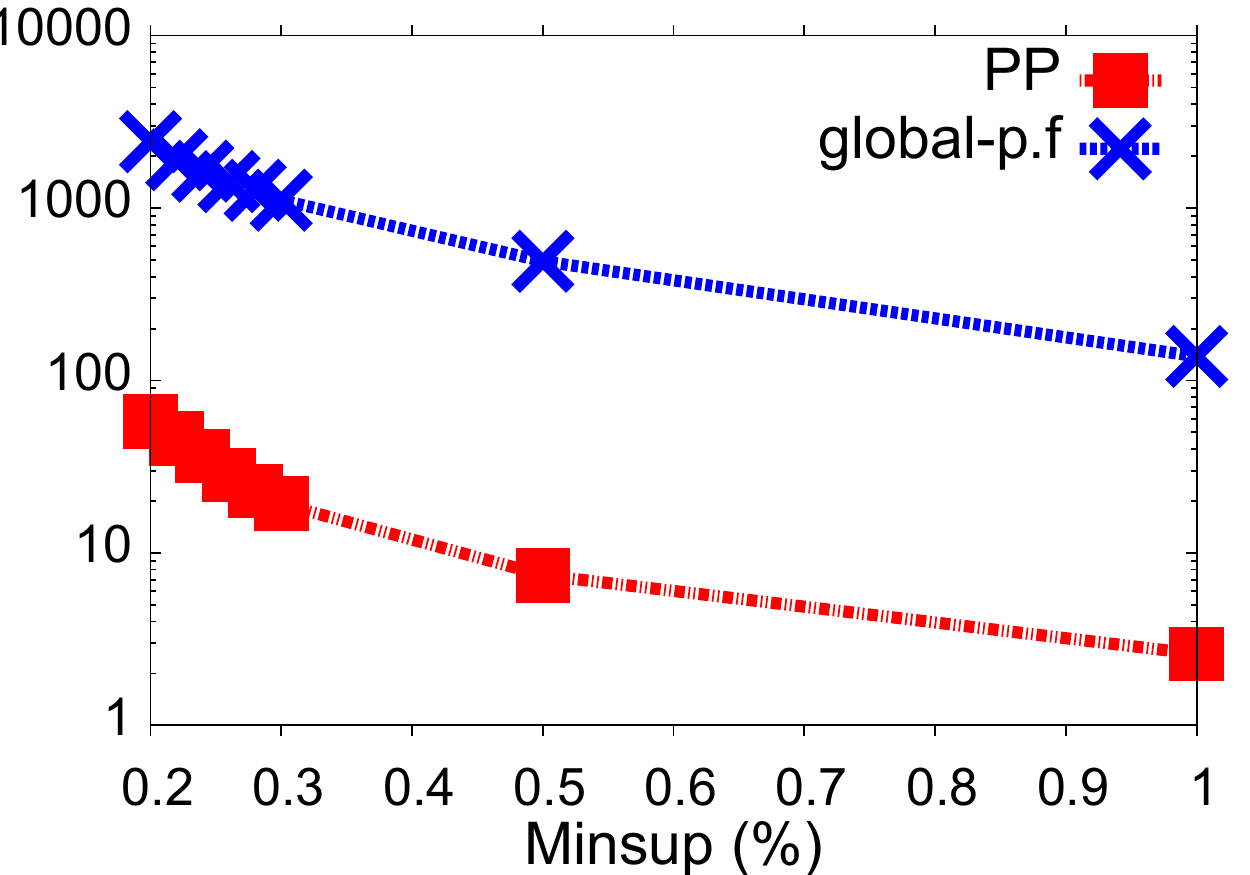} %Freq-Kosarak-col.pdf}
&
\includegraphics[width=3.8cm, height=2.8cm]{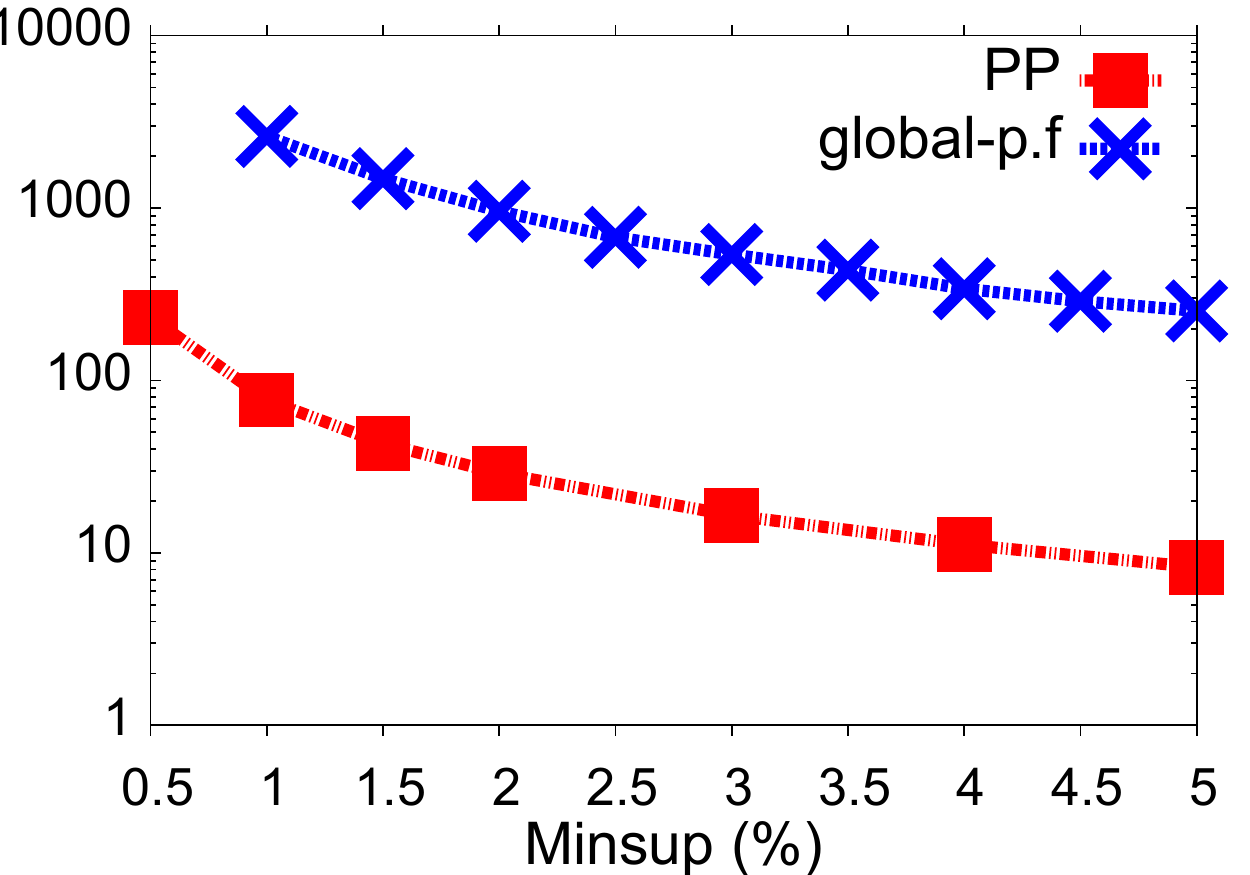} %Freq-FIFA-col.pdf}
\\
\includegraphics[width=3.8cm, height=2.8cm]{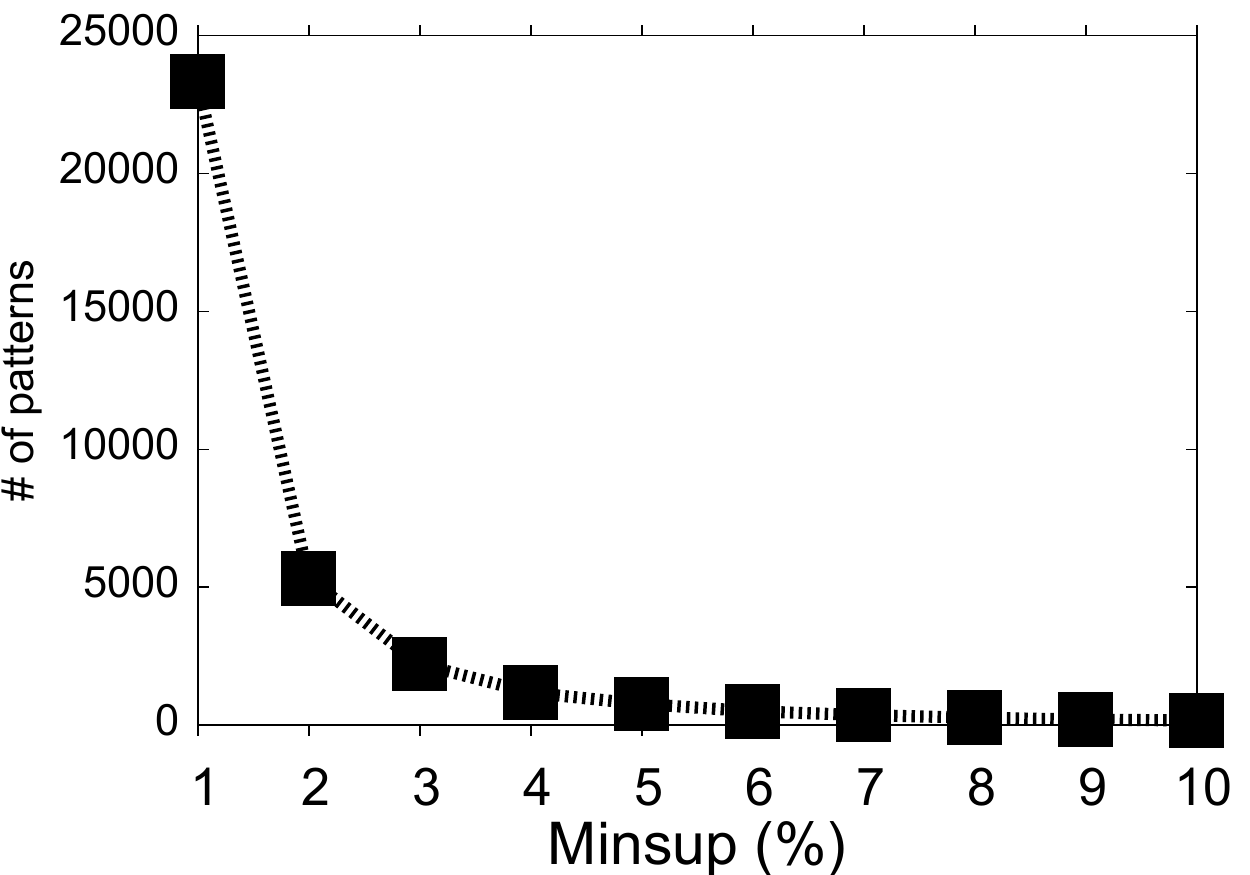} %Freq-BIBLE-col.pdf}     
&   
\includegraphics[width=3.8cm, height=2.8cm]{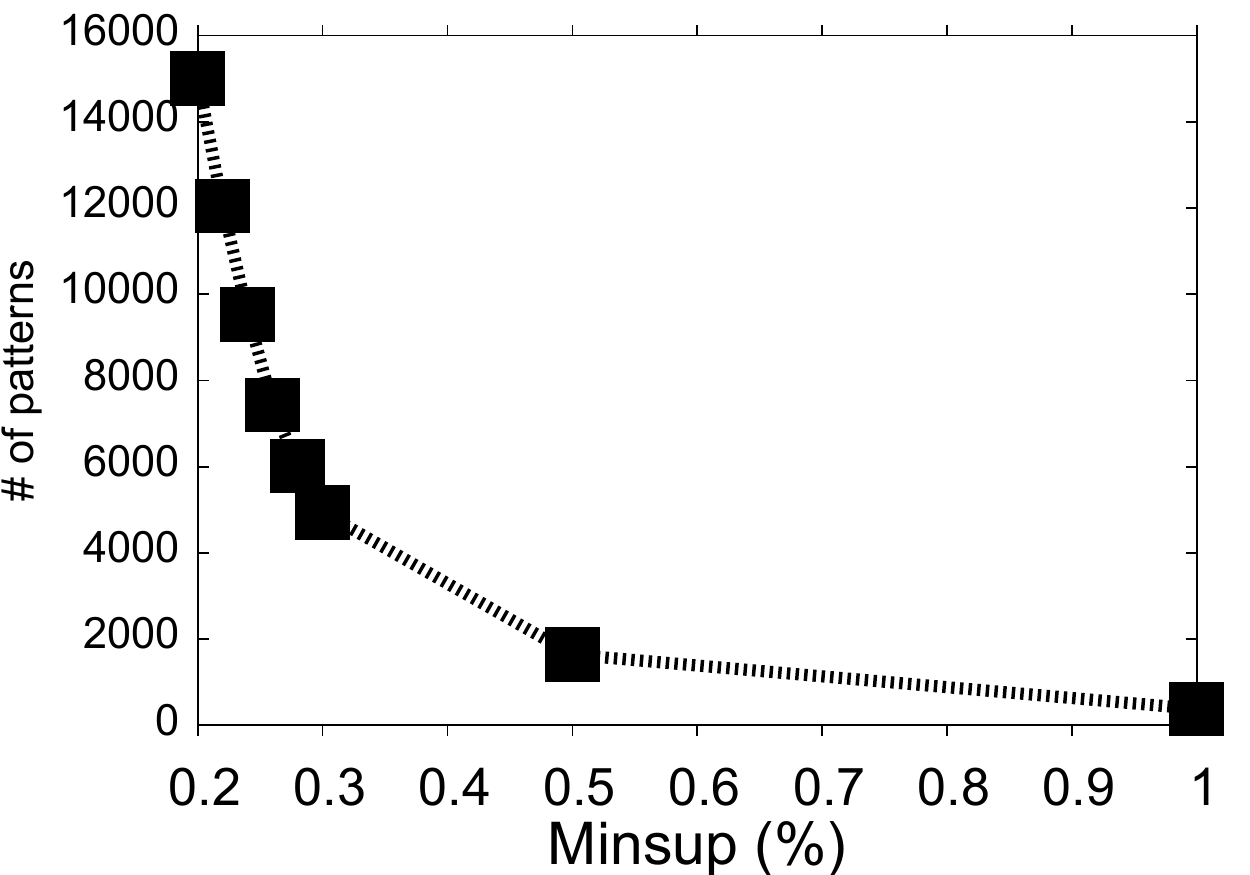} %Freq-Kosarak-col.pdf}
&
\includegraphics[width=3.8cm, height=2.8cm]{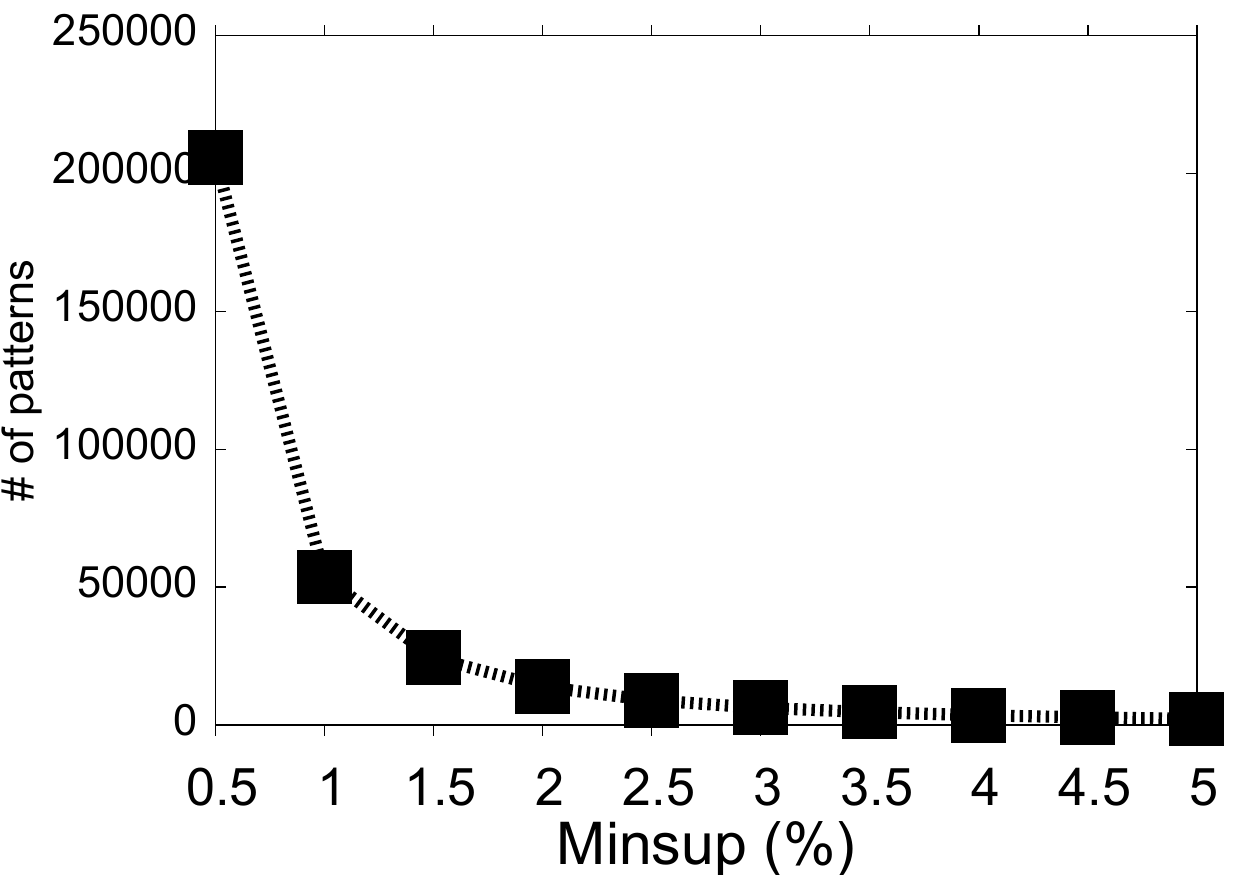} %Freq-FIFA-col.pdf}
\\
FIFA & Leviathan & Protein \\
\includegraphics[width=3.8cm, height=2.8cm]{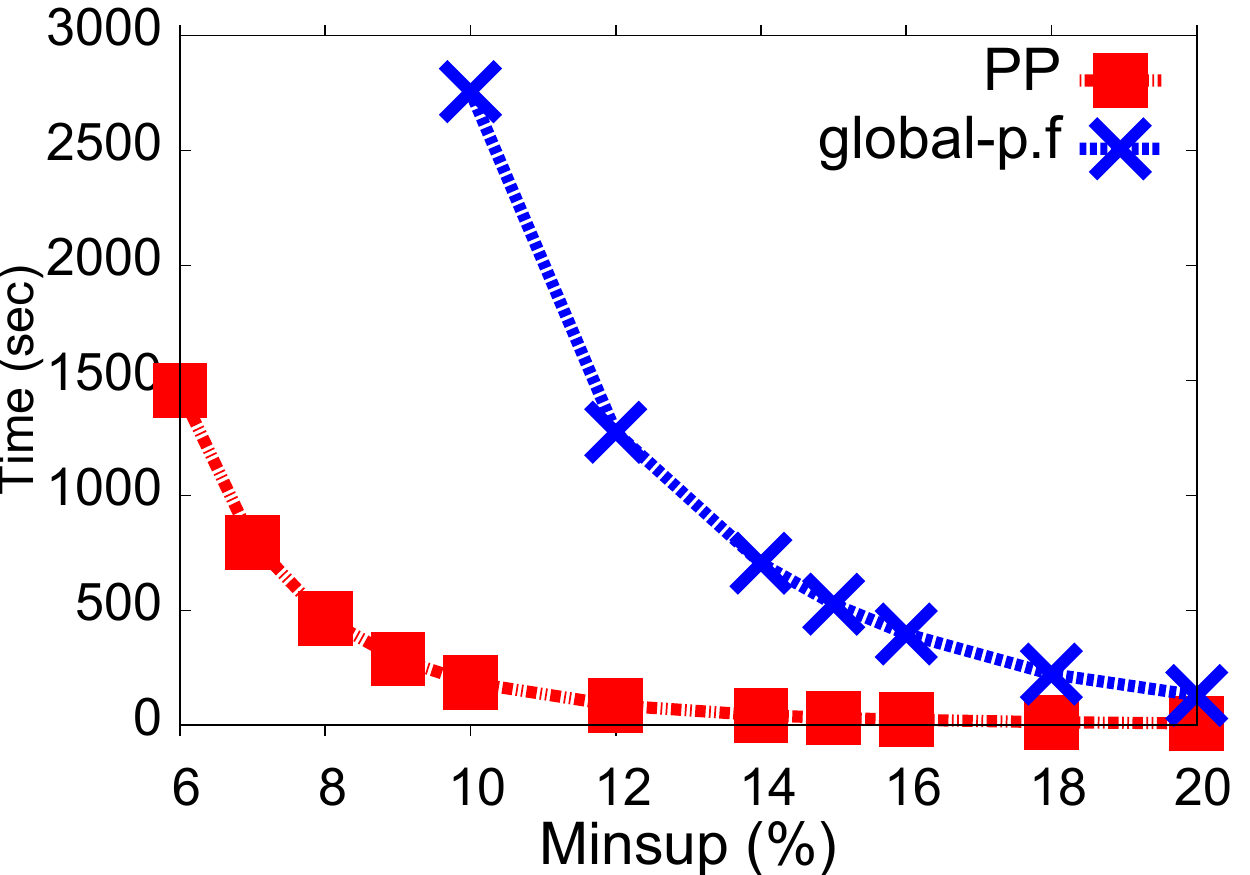} %Freq-FIFA-col.pdf}
&
\includegraphics[width=3.8cm, height=2.8cm]{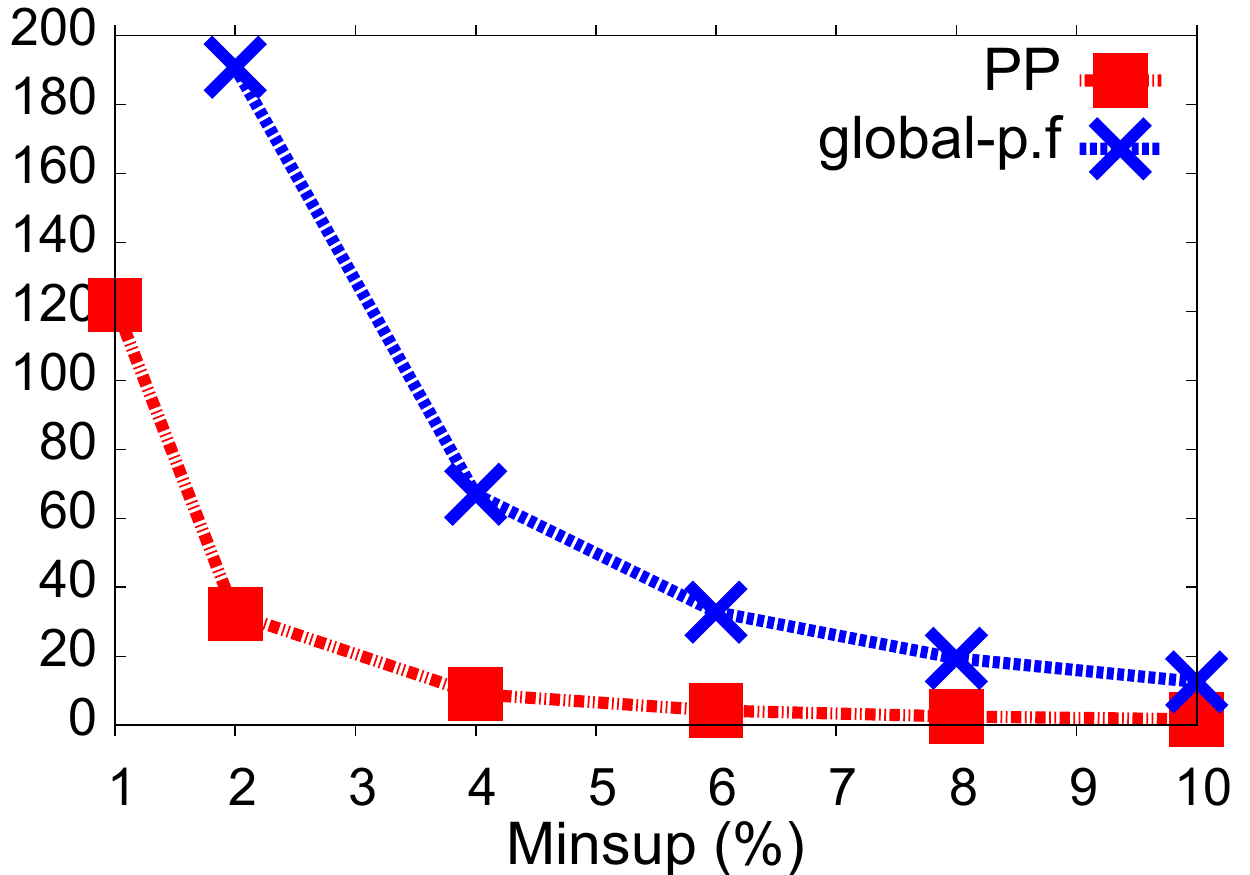} %Freq-Leviathan-col.pdf}
&
\includegraphics[width=3.8cm, height=2.8cm]{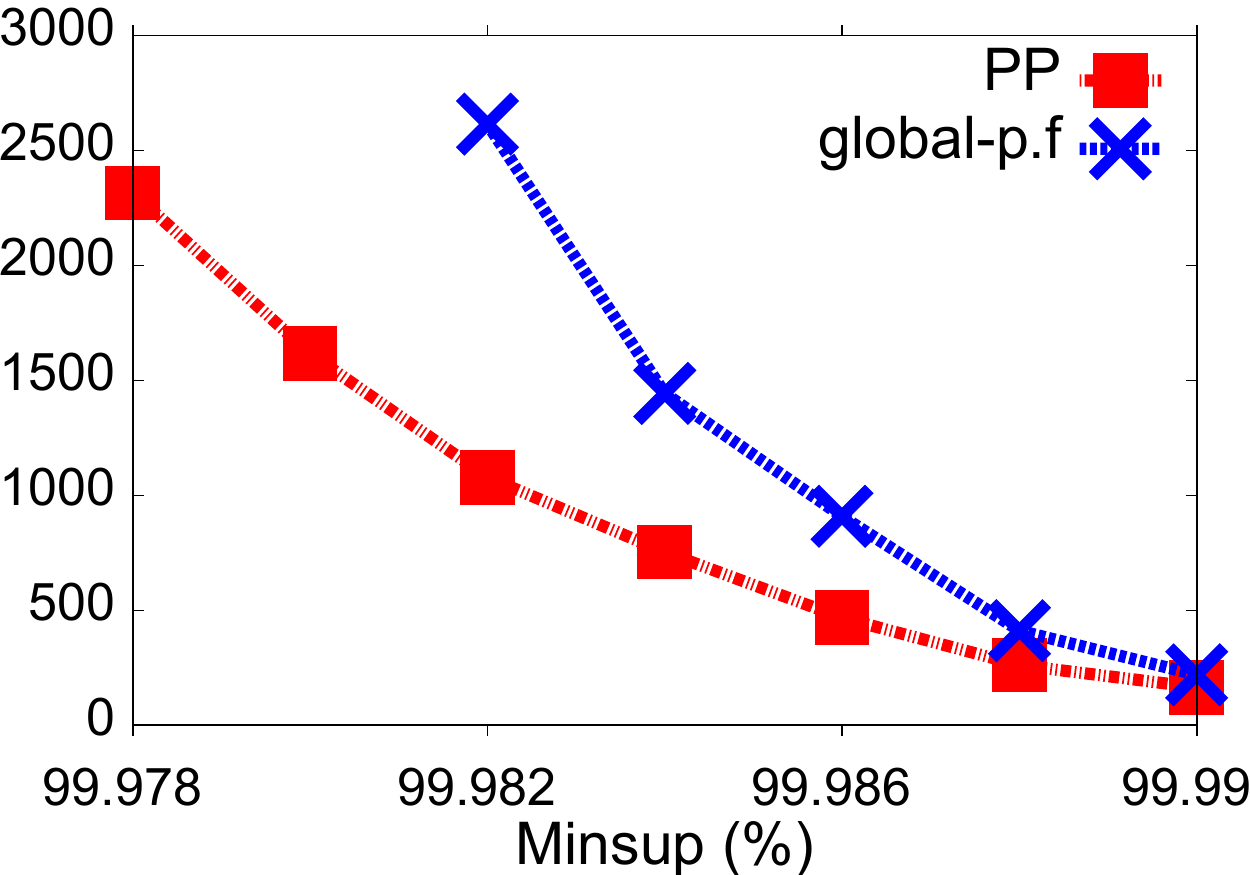} %Freq-FIFA-col.pdf}
\\
\includegraphics[width=3.8cm, height=2.8cm]{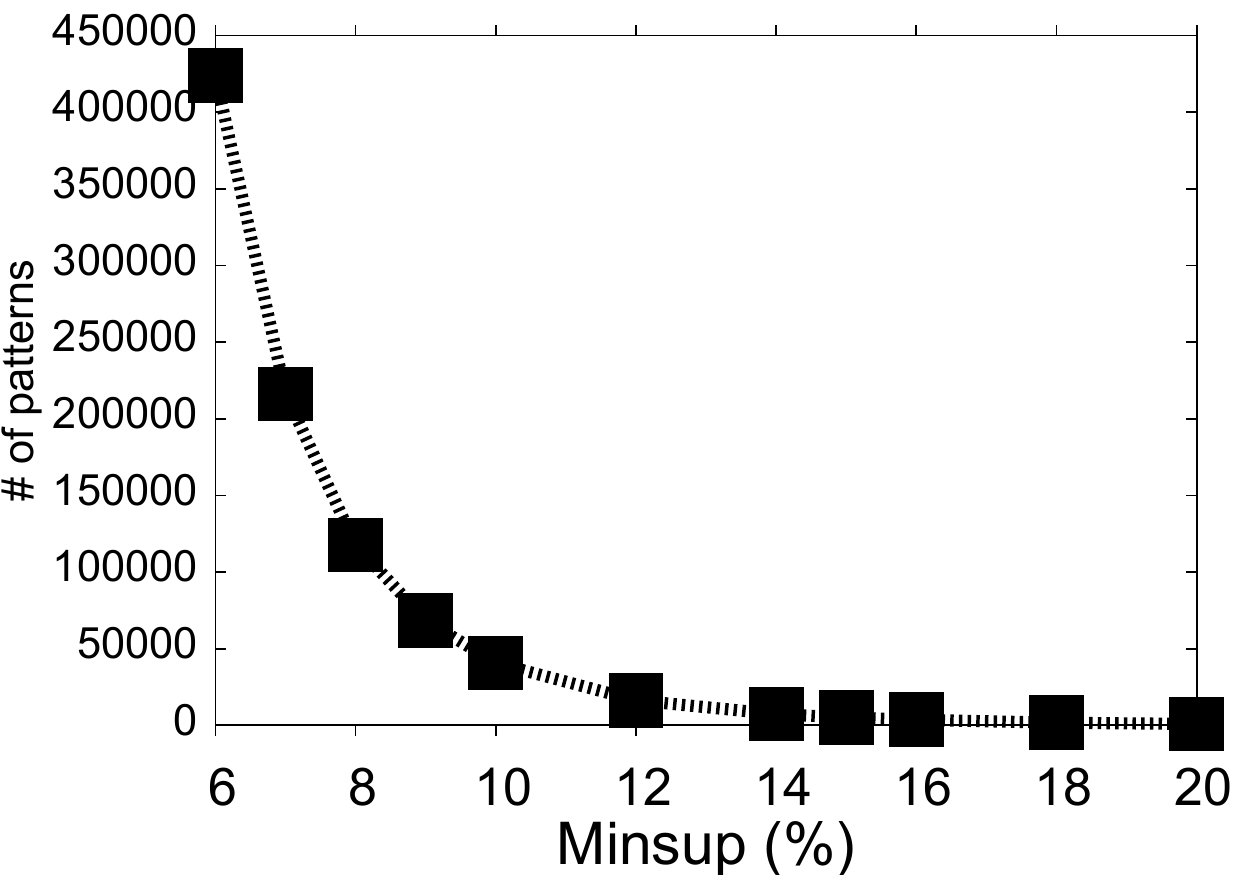} %Freq-FIFA-col.pdf}
&
\includegraphics[width=3.8cm, height=2.8cm]{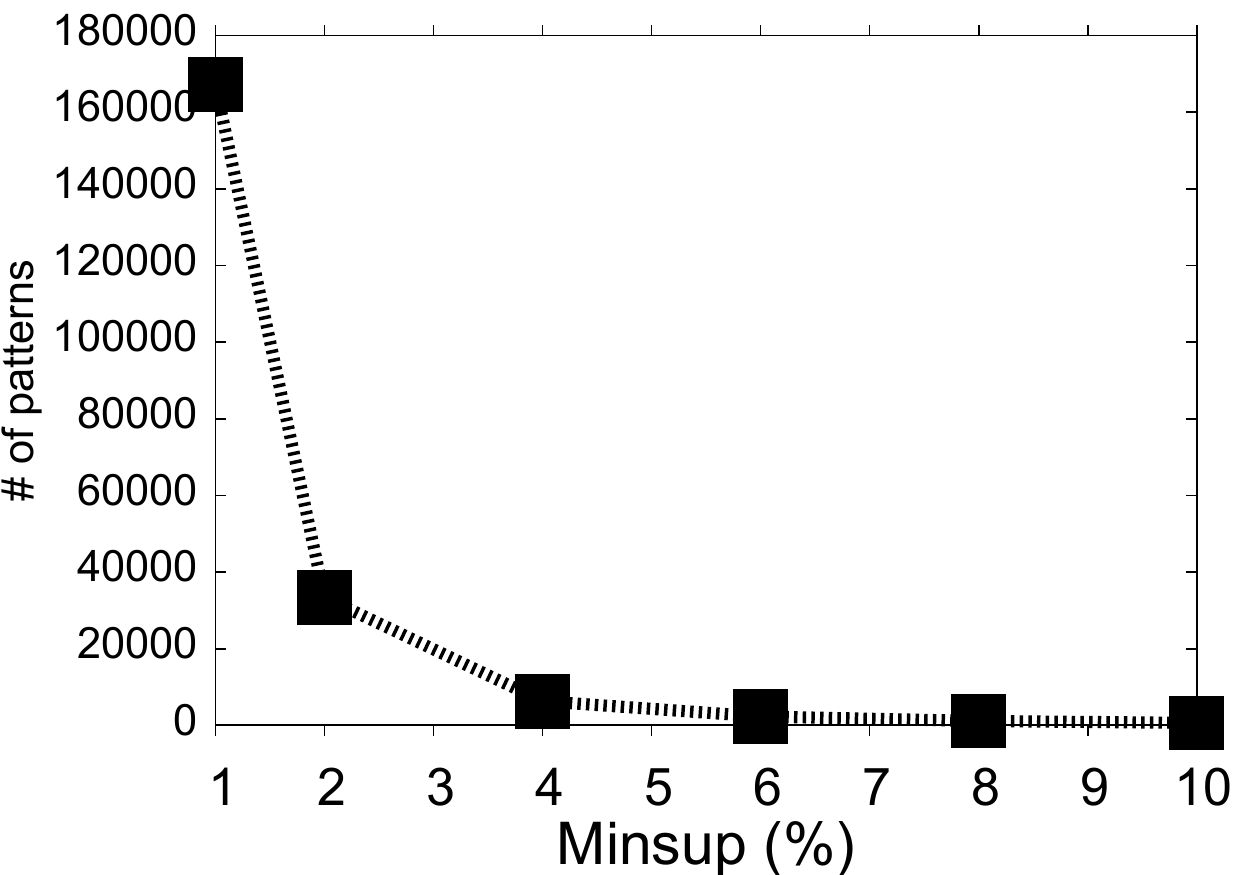} %Freq-Leviathan-col.pdf}
&
\includegraphics[width=3.8cm, height=2.8cm]{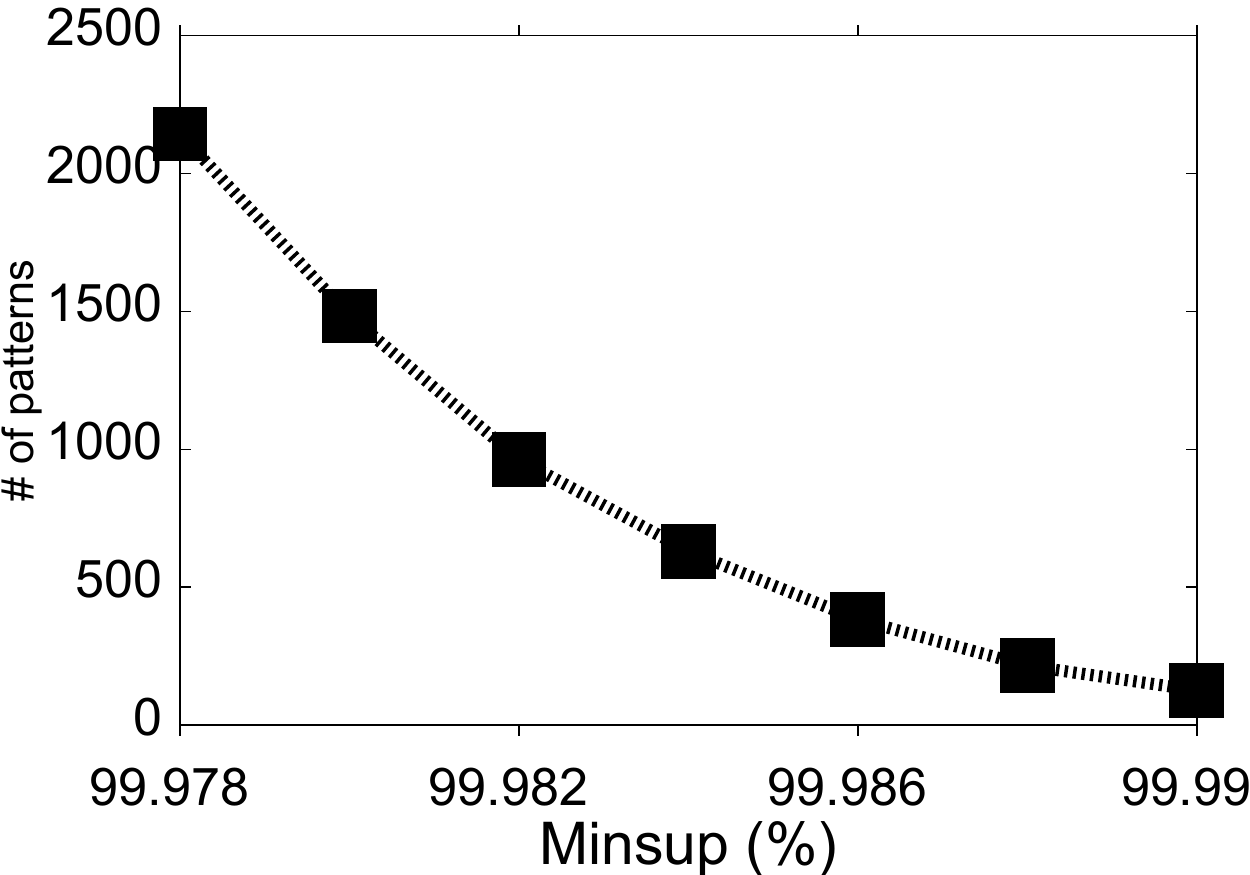} %Freq-FIFA-col.pdf}
\end{tabular}}
\vspace*{-.35cm}
 
\caption{\small \label{fig:FreqCPSM} Comparing \PP with \cpsm
  for SPM on real-life datasets: CPU times (top) and number of patterns (bottom).} 
\end{figure*}
%%%%%%%%%%%%%%%%%%%%%%%%%%%%%%%%%%%%%%%%%%%%%%%%%%%%%%%%%%%%%%%%%%%%%%%%%%%

\subsection{Comparing with CP Methods for SPM}
First we compare \PP with the two CP encodings 
  \cpsm and \cps (see Section
  \ref{CP4SPM}). 
Fig.~\ref{fig:FreqCPSM} shows the number of extracted sequential patterns and the CPU times
to extract them (in logscale for BIBLE, Kosarak and PubMed) for the
three methods. 

First, as expected, the lower $minsup$ is, the larger the number of
extracted sequential patterns. 
Second, when comparing the CPU times, \cps is the least performer
  method. On all the datasets, it fails to complete the extraction within the time 
  limit for all values of $minsup$ we considered. 
Third, \PP largely dominates \cpsm on all the datasets: 
\PP is more than an order of magnitude faster than \cpsm. 
The gains in terms of CPU times are greatly amplified for low values
of $minsup$. On BIBLE (resp. PubMed), the speed-up is
$84.4$ (resp. $33.5$) for $minsup$ equal to $1\%$. 
Another important observation that can be made is that, on most of the
datasets (except BIBLE and Kosarak), \cpsm is not able to mine for
patterns at very low frequency within the time limit. 
For example on FIFA, \PP is able to
complete the extraction for values of $minsup$ up to $6\%$ in $1,457$ 
seconds, while \cpsm fails to complete the extraction for $minsup$
less than $10\%$. The same trend is also conformed on Leviathan, where
\cpsm is not able to mine for patterns at $1\%$ minimum frequency.

To complement the results given by  Fig.~\ref{fig:FreqCPSM},
Table~\ref{table:SolverStat} reports for different datasets and different
values of $minsup$, the number of calls to the propagate routine of
{\tt Gecode} (column 5), and the number of nodes of the search tree (column
6). 
First, \PP explores less nodes than \cpsm. But, the difference is
not huge (gains of 45\%  and 33\% on FIFA and BIBLE
respectively). 
Second, our approach is very
effective in terms of number of propagations. For \PP, the number of 
propagations remains small (in thousands for small values of $minsup$)
compared to  \cpsm (in millions). 
This is due to 
the huge number of reified constraints used in \cpsm to encode the
subsequence relation. On the contrary, our \prefixCP global constraint
does not require any reified constraints nor any extra variables to
encode the subsequence relation. %To the best of our knwoledge, it is
%the first CP approach which aims to free from reified constraints. 

%%%%%%%%%%%%%%%%%%%%%%%%%%%%%%%%%%%%%%%%%%%%%%%%%%%%
\begin{table*}[t] \centering
\scalebox{0.85}{
\begin{tabular}{|l|l|l|r|r|r|r|r|r|}
\hline
\multirow{2}{*}{Dataset} & \multirow{2}{*}{$minsup$ (\%)} &
\multirow{2}{*}{\#PATTERNS} & \multicolumn{2}{c|}{CPU times (s)} & \multicolumn{2}{c|}{\#PROPAGATIONS} & \multicolumn{2}{c|}{\#NODES}\\
\cline{4-9}
&  &  & {\tt PP} & \cpsm & {\tt PP} & \cpsm & {\tt PP} & \cpsm \\
\hline
\multirow{6}{*}{FIFA} &20 & 938 & {\bf 8.16} & 129.54 & {\bf 1884} & 11649290 & {\bf 1025} & 1873 \\
&18 & 1743 &{\bf 13.39} & 222.68 & {\bf 3502} & 19736442  & {\bf 1922} & 3486 \\
&16 & 3578 & {\bf 24.39} & 396.11 &{\bf 7181} & 35942314 & {\bf 3923} & 7151\\
&14 & 7313 & {\bf 44.08} & 704 & {\bf 14691} & 65522076 & {\bf 8042} & 14616\\
&12 & 16323 & {\bf 86.46} & 1271.84 & {\bf 32820} & 126187396 & {\bf 18108} & 32604 \\
&10 & 40642 & {\bf 185.88} & 2761.47 & {\bf 81767 } & 266635050  & {\bf 45452} & 81181  \\
\hline
\hline
\multirow{6}{*}{BIBLE} &10 & 174 &{\bf 1.98}  & 105.01 & {\bf 363}  & 4189140& {\bf 235} & 348 \\
&8 & 274 & {\bf 2.47}  & 153.61 & {\bf 575}  & 5637671 & {\bf 362} & 548 \\
&6 & 508 & {\bf 3.45}  & 270.49 & {\bf 1065} & 8592858 & {\bf 669} & 1016 \\
&4 & 1185 & {\bf 5.7}  & 552.62 & {\bf 2482} & 15379396 & {\bf 1575}  & 2371 \\
&2 & 5311 & {\bf 15.05}  & 1470.45 & {\bf 11104}  & 39797508  &  {\bf 7048} & 10605  \\ 
&1 & 23340 & {\bf 41.4} & 3494.27 & {\bf 49057}  & 98676120 & {\bf 31283} & 46557 \\
\hline
\hline
\multirow{6}{*}{PubMed} &5 &  2312 & {\bf 8.26} &  253.16 & {\bf 4736}   & 15521327  & {\bf 2833}  &  4619 \\ 
&4 & 3625 & {\bf 11.17}  & 340.24 &{\bf 7413} & 20643992 & {\bf 4428}  & 7242 \\
&3 & 6336 & {\bf 16.51} & 536.96 &{\bf 12988} &29940327  & {\bf 7757} & 12643 \\
&2 & 13998 & {\bf 28.91} & 955.54 & {\bf 28680} &50353208  & {\bf 17145} & 27910  \\
&1 & 53818 & {\bf 77.01} & 2581.51 & {\bf 110133} &  124197857 &  {\bf 65587} &  107051 \\
\hline
\hline
\multirow{6}{*}{Protein} &99.99 & 127 &{\bf 165.31} & 219.69 & {\bf 264} & 26731250 & {\bf 172} & 221 \\
&99.988 & 216 & {\bf 262.12} & 411.83 & {\bf 451} & 44575117& {\bf 293} & 390\\
&99.986 & 384 & {\bf 467.9}6 & 909.47& {\bf 805} &80859312 & {\bf 514} & 679\\
&99.984 & 631 & {\bf 753.3} & 1443.92 & {\bf 1322} & 132238827 & {\bf 845} & 1119\\
&99.982 & 964& {\bf 1078.73} & 2615 & {\bf 2014} & 201616651 &  {\bf 1284} & 1749\\ 
&99.98 & 2143 & {\bf 2315.65} & {\bf $-$} & {\bf 4485}  & {\bf $-$} & {\bf 2890} & {\bf $-$} \\
\hline
\hline
\multirow{6}{*}{Kosarak} &1 & 384 &{\bf 2.59} & 137.95 & {\bf 793} & 8741452 & {\bf 482} & 769\\
&0.5 & 1638 & {\bf 7.42} & 491.11 & {\bf 3350} & 26604840 & {\bf 2087} & 3271\\
&0.3 & 4943& {\bf 19.25} & 1111.16& {\bf 10103} & 56854431& {\bf 6407} &  9836\\
&0.28 & 6015 & {\bf 22.83} & 1266.39 & {\bf 12308} & 64003092 & {\bf 7831} & 11954\\
&0.24 & 9534 & {\bf 36.54} & 1635.38 & {\bf 19552} & 81485031 &  {\bf 12667} & 18966 \\
&0.2  & 15010& {\bf 57.6} & 2428.23 & {\bf 30893} & 111655799 &  {\bf 20055} & 29713 \\
\hline
\hline
\multirow{6}{*}{Leviathan} &10 & 651 &{\bf 1.78} & 12.56 & {\bf 1366} &  2142870& {\bf 849} & 1301\\
&8 & 1133 & {\bf 2.57} & 19.44 & {\bf 2379} &3169615  & {\bf 1487} & 2261\\
&6 & 2300 & {\bf 4.27} &32.85 & {\bf 4824} &5212113 & {\bf 3008} &4575 \\
&4 & 6286 & {\bf 9.08} &  66.31& {\bf 13197} &10569654  & {\bf 8227} &12500  \\
&2 & 33387& {\bf 32.27} & 190.45& {\bf 70016} &  33832141&  {\bf 43588} &66116 \\ 
&1 & 167189& {\bf 121.89} & $-$ & {\bf 350310} & $-$ &  {\bf 217904} & $-$\\ 
\hline
\end{tabular}
}
%\vspace{-0.35cm}
\caption{\PP vs. \cpsm.} 
\label{table:SolverStat}
\end{table*}
%%%%%%%%%%%%%%%%%%%%%%%%%%%%%%%%%%%%%%%%%%%%%%%%%%%%
\begin{figure*}[t]
{\footnotesize
\begin{tabular}{ccc}
BIBLE & Kosarak & PubMed   \\
\includegraphics[width=4cm, height=3.0cm]{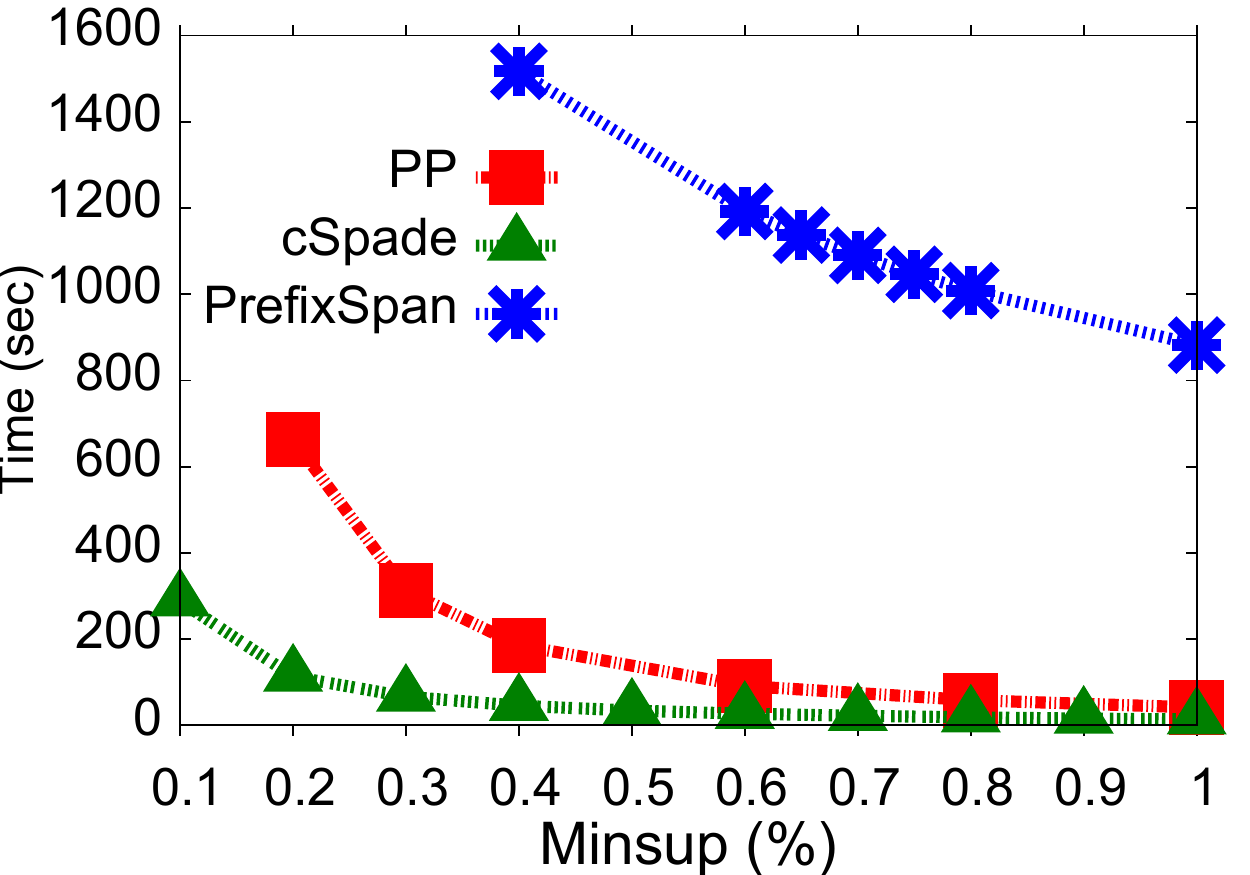}%BIBLEC.pdf} %Freq-BIBLE-col.pdf}     
&   
\includegraphics[width=4cm, height=3.0cm]{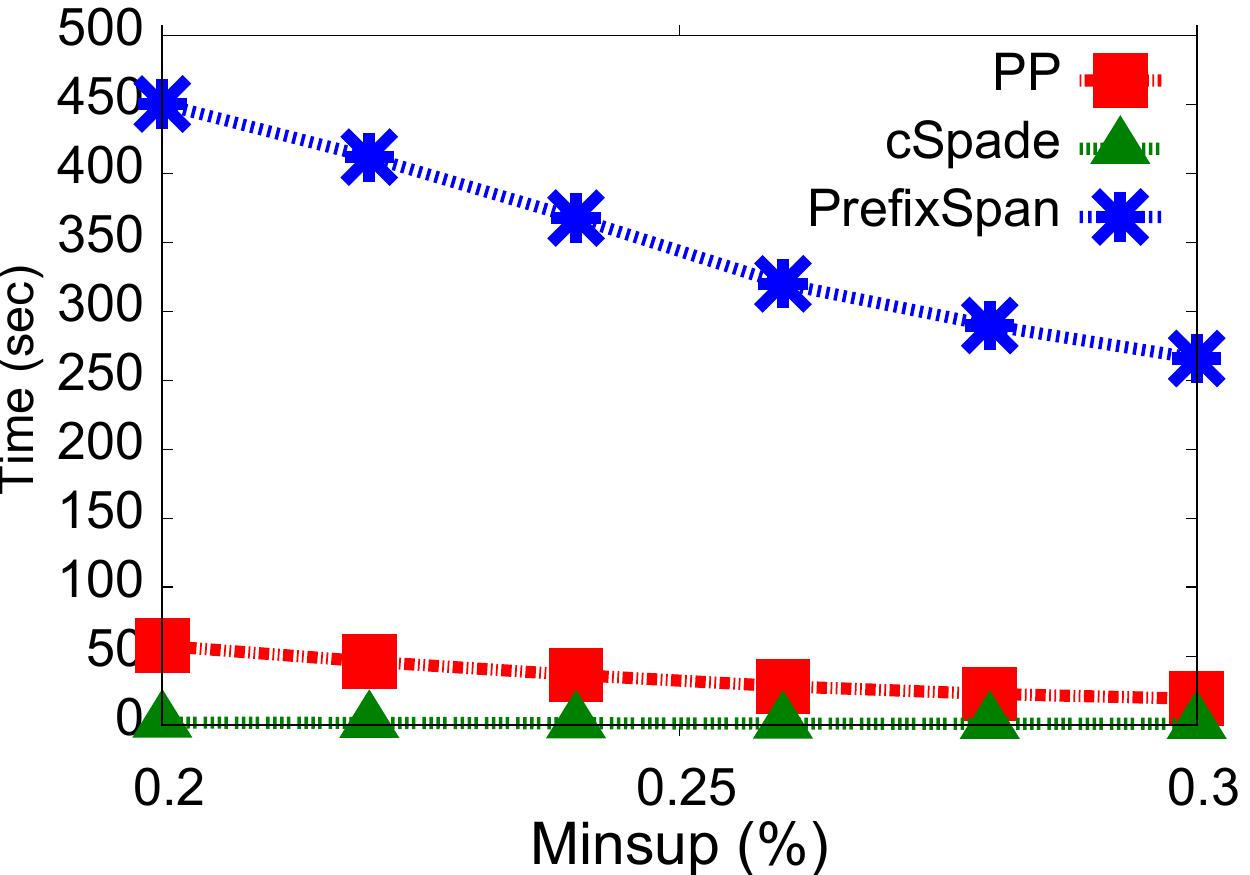}%KosarakC.pdf} %Freq-Kosarak-col.pdf}
&
\includegraphics[width=4cm, height=3.0cm]{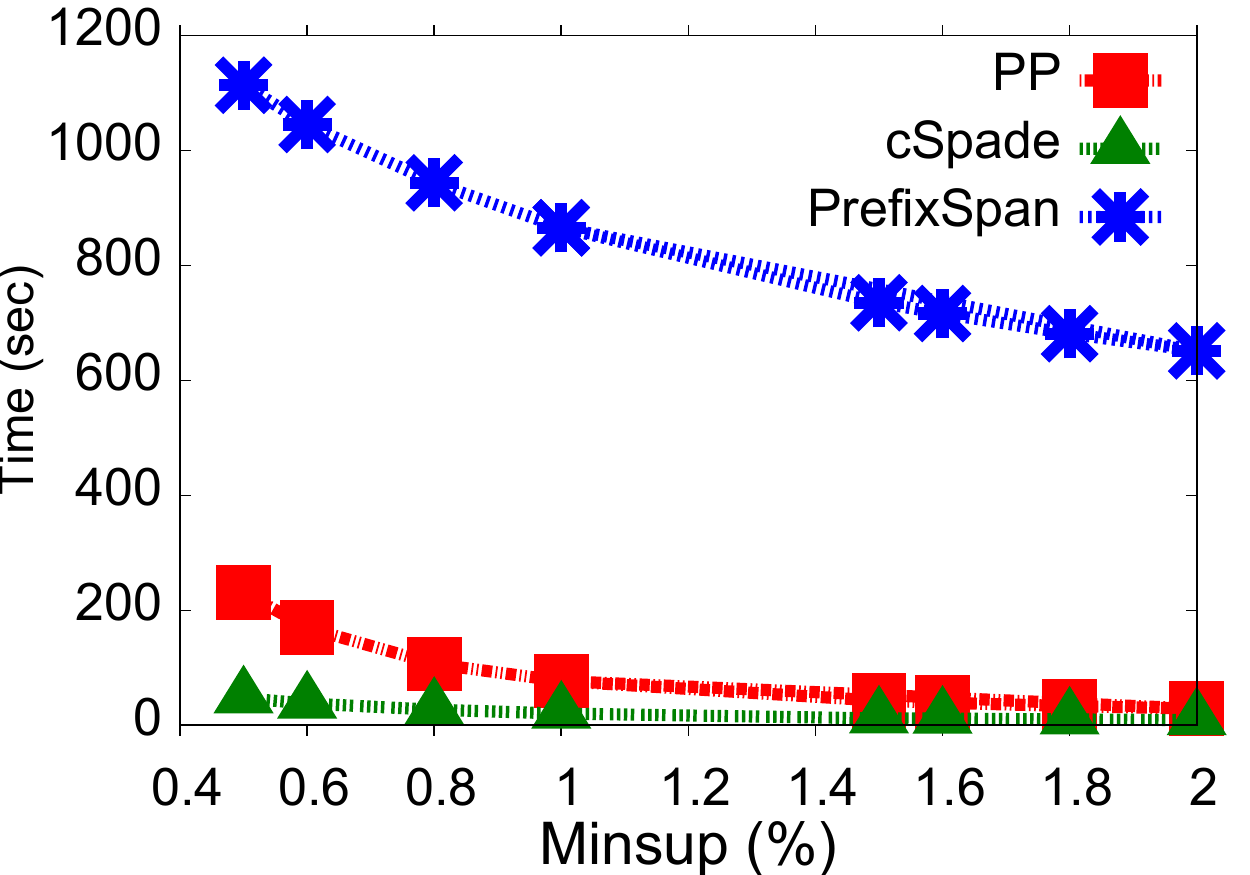}%Freq-pubmed-col.pdf}    
\\
FIFA & Leviathan & Protein \\
\includegraphics[width=4cm, height=3.0cm]{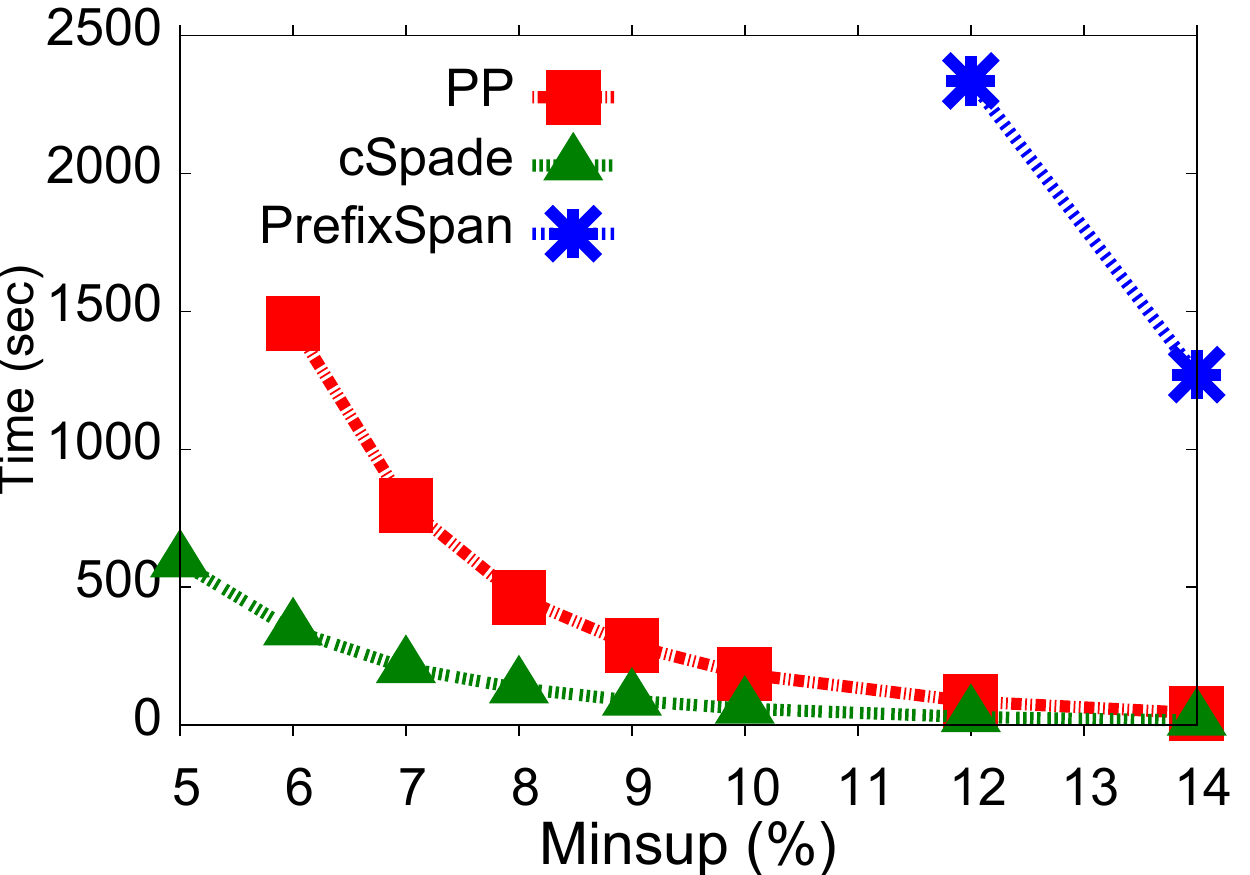}%FIFAC.pdf} %Freq-FIFA-col.pdf}
&
\includegraphics[width=4cm, height=3.0cm]{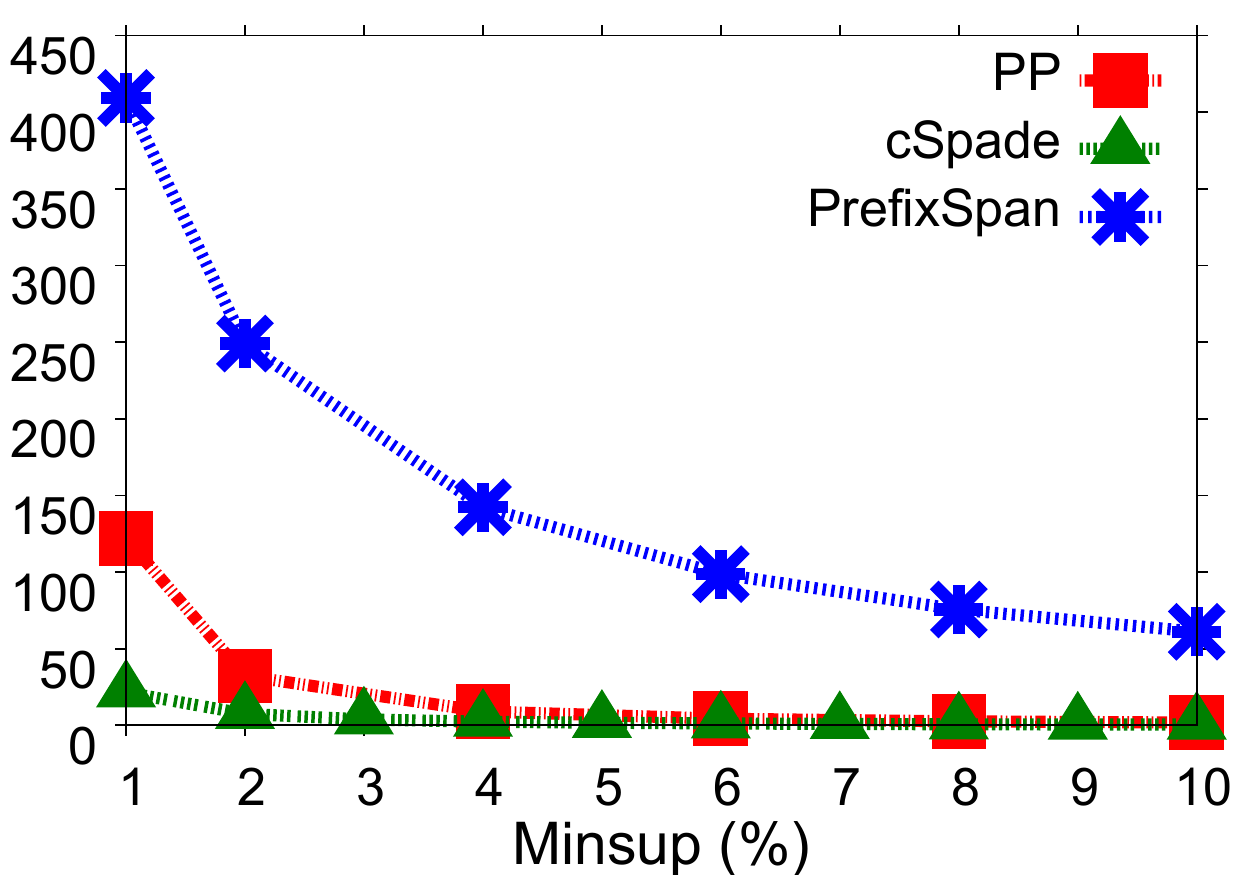}%LeviathanC.pdf} %Freq-Leviathan-col.pdf}
&
\includegraphics[width=4cm, height=3.0cm]{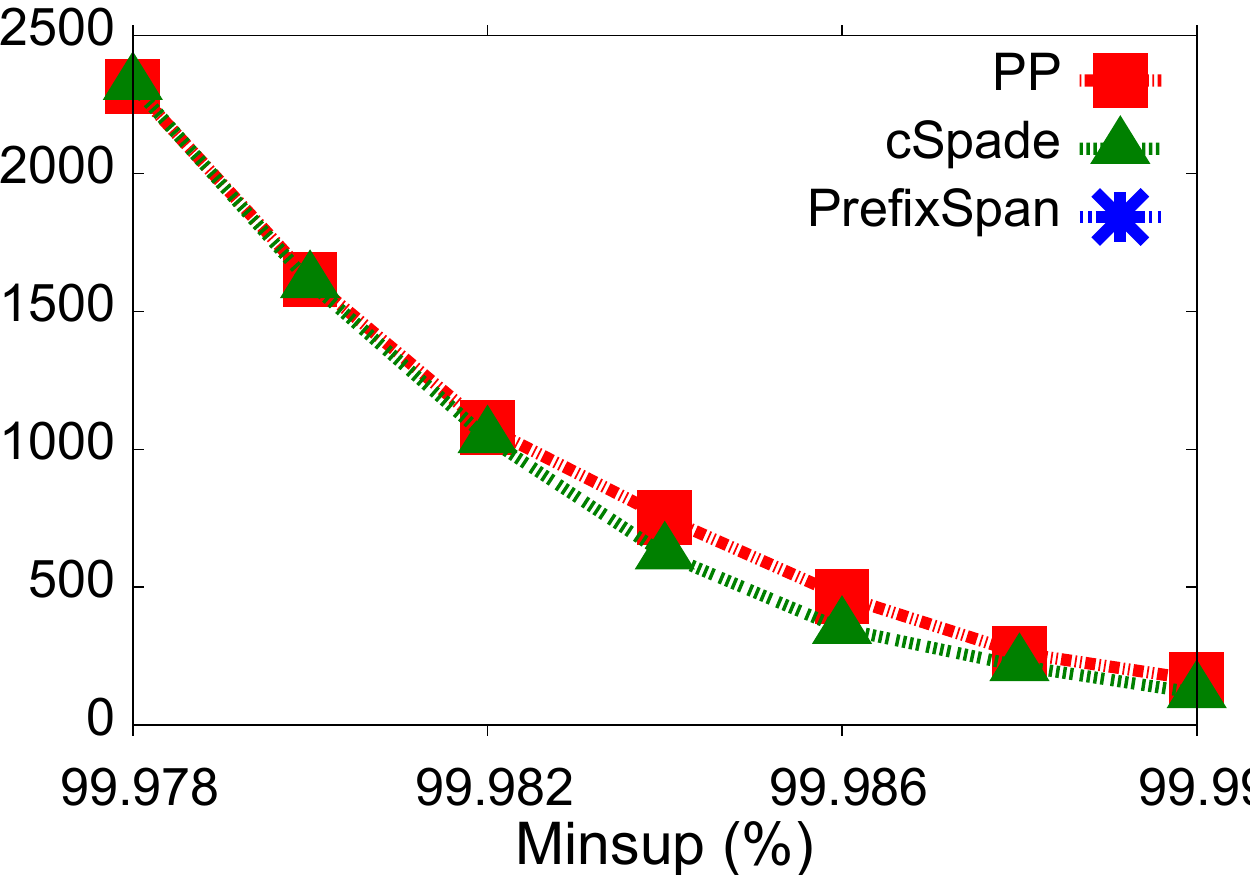}
\vspace*{-.35cm}
\end{tabular}}

\caption{\small \label{fig:FreqCLASSIC} Comparing \prefixCP with
  state-of-the-art algorithms for SPM.} 
\end{figure*}
%%%%%%%%%%%%%%%%%%%%%%%%%%%%%%%%%%%%%%%%%%%%%%%%%%%%

\subsection{Comparing with ad hoc Methods for SPM}
Our second experiment compares \PP with state-of-the-art methods  
for SPM. Fig.~\ref{fig:FreqCLASSIC} shows the CPU times of the three 
methods.
First, \cspade obtains the best
performance on all datasets (except on Protein).  
However, \PP exhibits a similar behavior as \cspade, but it is less
faster (not counting the highest values of $minsup$).  
The behavior of \cspade on Protein is due to the
  vertical representation format that is not appropriated in the
  case of databases having large sequences and small number of
  distinct items, thus degrading the performance of the mining
  process.
Second, \PP which also uses the concept of projected databases,
clearly outperforms \prefix on all datasets. 
This is due to
our filtering algorithm combined together with incremental data
structures to manage the projected databases.  
On FIFA, \prefix is not
able to complete the extraction for $minsup$ less than $12\%$, while
our approach remains feasible until $6\%$ within the time limit. 
On Protein, \prefix fails to complete the
  extraction for all values of $minsup$ we considered. 
%%
%{\color{blue}Finally, we observe that the performance of \cspade is decreased for the Protein dataset. This is explained by the huge number of sequences in this dataset with a small number of distinct items~\footnote{Dense datasets are characterized by having a small number of distinct items and a large number of sequences.}. For the same dataset, \prefix doesn't succeed to terminate the extraction within the time limit (it doesn't appear in the figure). In the other hand, our approach keeps the same behaviour obtained for the other datasets: it get the same performance as \cspade.}
%
These results clearly demonstrate that our approach competes well with
state-of-the-art methods for SPM on large datasets and achieves scalability 
while it is a major issue of existing CP approaches. 

\subsection{SPM under size and item constraints}
Our third experiment aims at assessing the interest of pushing
simultaneously different types of constraints. We impose on the PubMed
dataset usual constraints such as {\it the minimum frequency} and the
{\it minimum size} constraints and other useful constraints expressing
some linguistic knowledge such as {\it the item
  constraint}. The goal is to retain sequential patterns which convey
linguistic regularities (e.g., gene - rare disease relationships)~\cite{BCCC2012cbms}.  
{\it The size constraint} allows to
remove patterns that are too small w.r.t. the
number of items (number of words) to be relevant patterns. We tested
this constraint with $\ell_{min}$ set to 3.  
{\it The item constraint} imposes that the extracted patterns must
contain the item GENE and the item DISEASE. 
As no ad hoc method exists for this combination of 
constraints, we only compare \PP with
\cpsm. 
Fig.~\ref{fig:const} shows the CPU times and the number of
sequential patterns extracted 
with and without constraints. 
First, pushing simultaneously the two constraints enables to reduce
significantly the number of patterns. Moreover, the CPU times for \PP decrease 
slightly whereas for \cpsm (with and without
constraints), they are almost the same. 
This
is probably due to the weak communication between
the $m$ \texttt{exists-embedding} reified global constraints and the
two constraints. This reduces significantly the quality of the
whole filtering. 
Second (see Table~\ref{table:const:stat}), when considering the
two constraints, \PP clearly dominates \cpsm (speed-up value up to
$51.5$). Moreover, the number of propagations performed by \PP remains 
very small as compared to \cpsm. 
Fig.~\ref{fig:const:prop} compares the two methods under 
the minimum size constraint for different values of 
$\ell_{min}$, with $minsup$ fixed to
$1\%$. Table~\ref{table:const:stat} compares the two methods in terms
of numbers of propagations (column $5$) and number of nodes of the
search tree (column $6$). 
Once again, \PP is always the most performer method (speed-up
value up to $53.1$). These results also confirm what we observed previously,
namely the weak communication between reified global constraints 
and constraints imposed on patterns (i.e., size and item constraints). 

%%%%%%%%%%%%%%%%%%%%%%%%%%%%%%%%%%%%%%%%%%%%%%%%%%%%%%%%%
\begin{figure*}[t]
\centering
\subfloat[\# of patterns\label{fig:const:patterns}]{
\includegraphics[width=4cm, height=2.6cm]{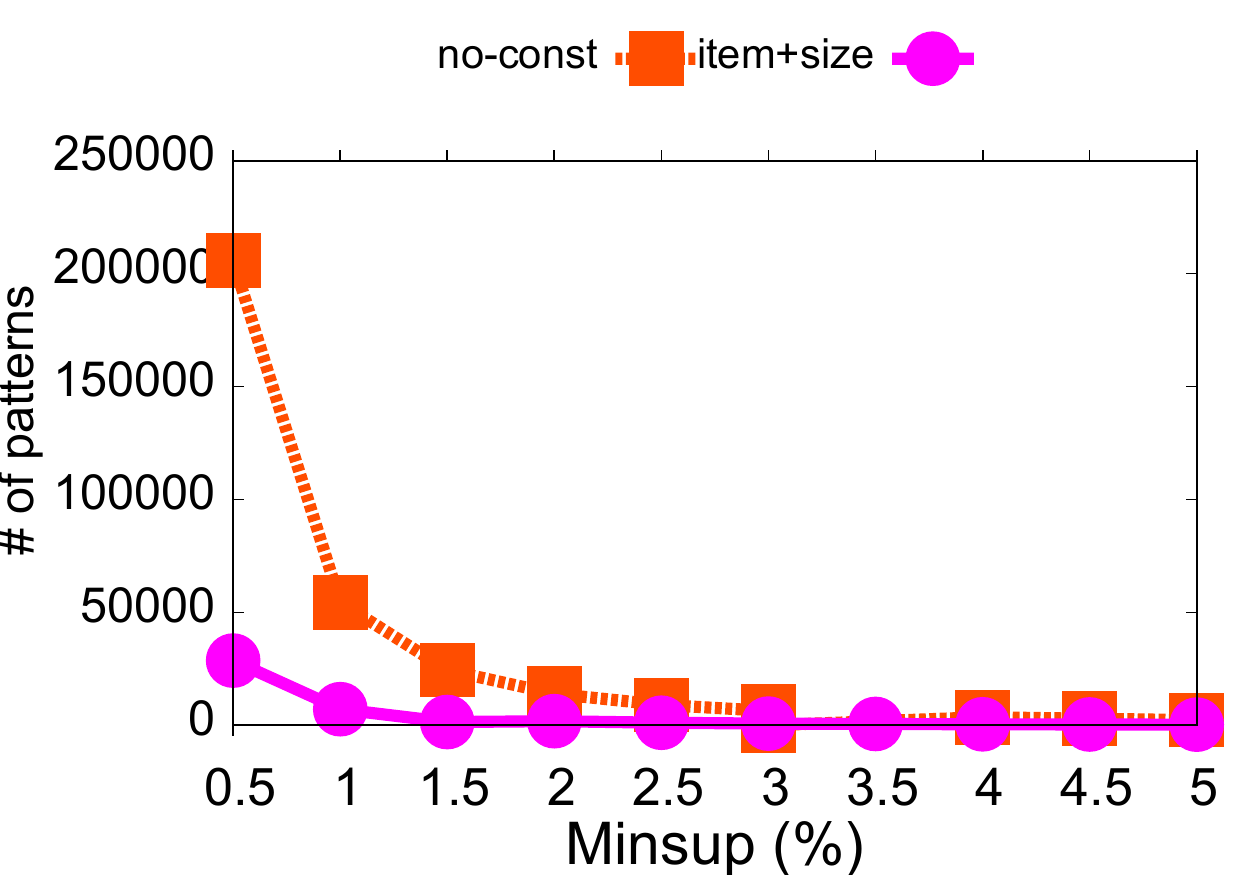}
}
\subfloat[CPU times (logscale) \label{fig:const:time}]{
\includegraphics[width=4cm, height=2.8cm]{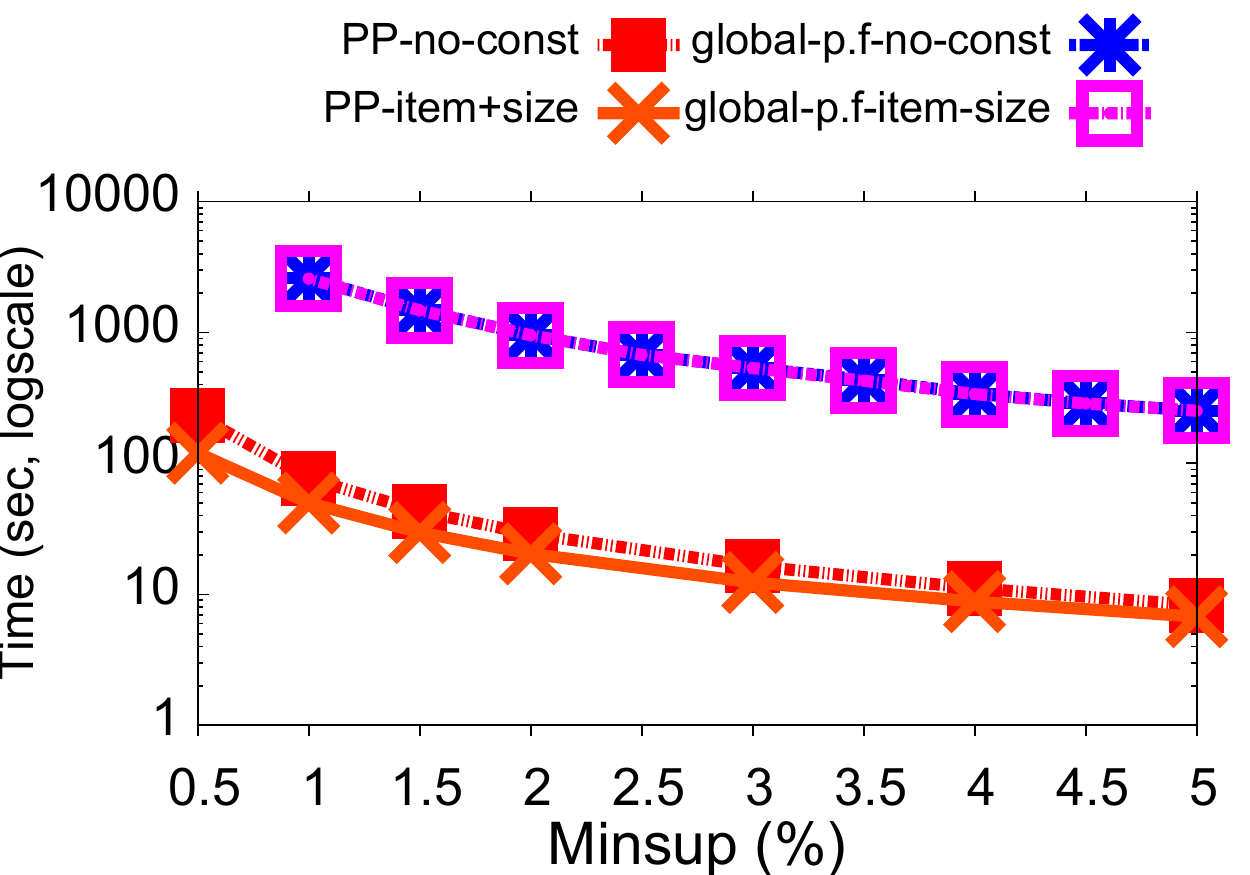}
}
\subfloat[Minimum size constraint\label{fig:const:prop}]{
\includegraphics[width=4cm, height=2.6cm]{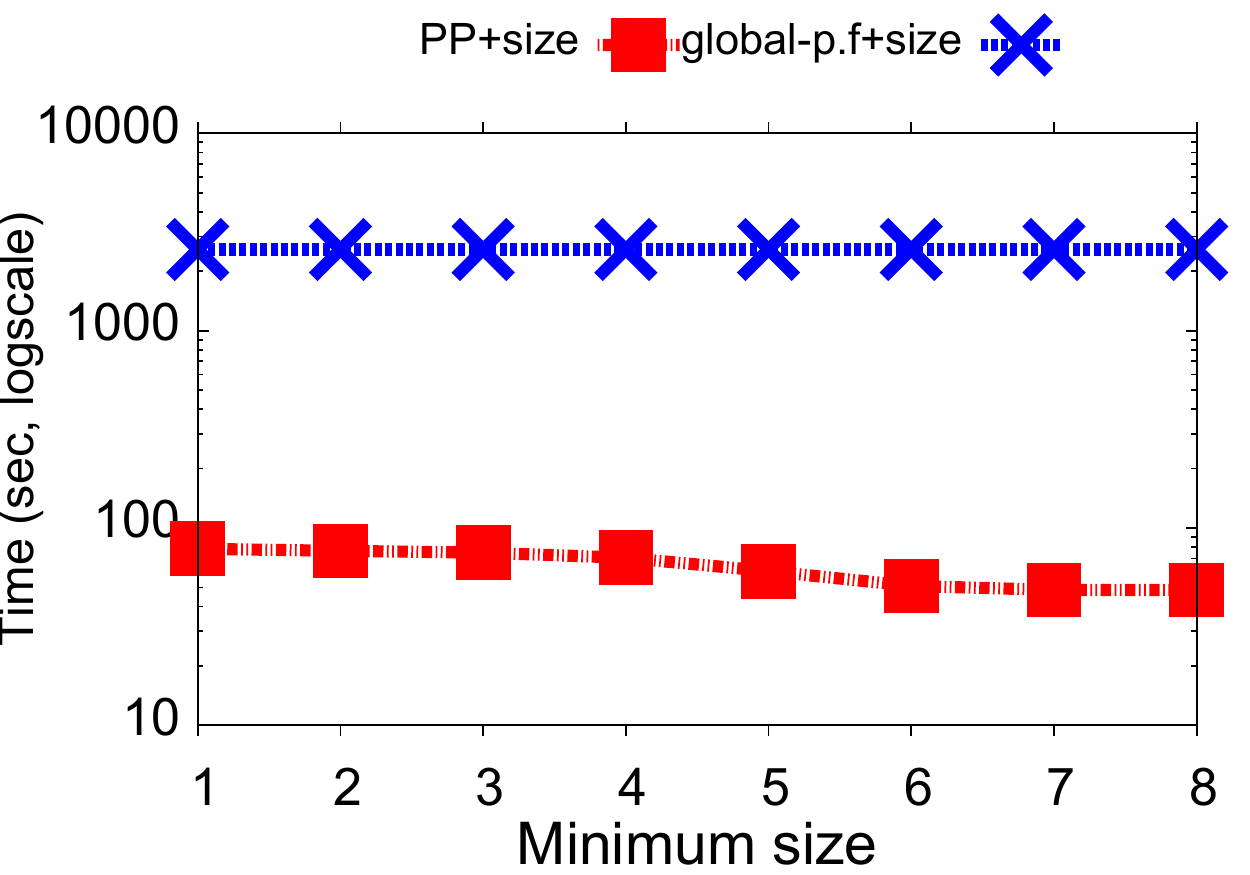}
}
\vspace*{-.35cm} 
\caption{\small Comparing \PP with \cpsm under minimum size and item constraints on PubMed.} \label{fig:const}
\end{figure*}
%%%%%%%%%%%%%%%%%%%%%%%%%%%%%%%%%%%%%%%%%%%%%%%%%%%%%%%%%
\begin{table*}[t] \centering
\scalebox{0.85}{
\begin{tabular}{|l|l|l|r|r|r|r|r|r|}
\hline
\multirow{2}{*}{Dataset} & \multirow{2}{*}{$minsup$ (\%)} &
\multirow{2}{*}{\#PATTERNS} & \multicolumn{2}{c|}{CPU times (s)} & \multicolumn{2}{c|}{\#PROPAGATIONS} & \multicolumn{2}{c|}{\#NODES}\\
\cline{4-9}
&  &  & {\tt PP} & \cpsm & {\tt PP} & \cpsm & {\tt PP} & \cpsm \\
\hline
\multirow{6}{*}{PubMed} & 5 & 279 & {\bf 6.76} & 252.36 & {\bf 7878} & 12234292  & {\bf 2285} & 4619  \\ 
&4 & 445 & {\bf 8.81}  & 339.09 & {\bf 12091} & 16475953 & {\bf 3618}  & 7242 \\
&3 & 799 & {\bf 12.35} & 535.32 & {\bf 20268} & 24380096 & {\bf 6271}   & 12643 \\
&2& 1837 & {\bf 20.41} & 953.32 & {\bf 43088}   &  42055022 & {\bf 13888}  & 27910  \\ 
&1 & 7187 & {\bf 49.98} & 2574.42 & {\bf 157899}  &  107978568 & {\bf 52508}  &  107051 \\
\hline
\end{tabular}
}
%\vspace{-0.35cm}
\caption{\small {\tt PP} vs. \cpsm under minimum size and item constraints.} 
\label{table:const:stat}
\end{table*}
%%%%%%%%%%%%%%%%%%%%%%%%%%%%%%%%%%%%%%%%%%%%%%%%%%%%

\begin{table*}[t] \centering
\scalebox{0.85}{
\begin{tabular}{|l|l|l|r|r|r|r|r|r|}
\hline
\multirow{2}{*}{Dataset} & \multirow{2}{*}{$\ell_{min}$} &
\multirow{2}{*}{\#PATTERNS} & \multicolumn{2}{c|}{CPU times (s)} & \multicolumn{2}{c|}{\#PROPAGATIONS} & \multicolumn{2}{c|}{\#NODES}\\
\cline{4-9}
&  &  & {\tt PP} & \cpsm & {\tt PP} & \cpsm & {\tt PP} & \cpsm \\
\hline
\multirow{6}{*}{PubMed} & 8 & 12 & {\bf 48.52} & 2577.09 & {\bf 55523} & 105343528  & {\bf 50264} &  107051\\ 
& 6 & 3596 & {\bf 50.91} & 2576.9 & {\bf 59144} & 106272419  & {\bf 50486} & 107051 \\ 
& 4 & 40669 & {\bf 70.61} & 2579.3 & {\bf 96871} & 117781215  & {\bf 59194} &  107051 \\ 
& 2 & 53486 & {\bf 76.64} & 2580.41  & {\bf 109801} &  123913176 & {\bf 65334} & 107051 \\ 
& 1 & 53818 & {\bf 78.49} & 2579.85  & {\bf 110133} &  117208559 & {\bf 65587} &  107051\\ 
\hline
\end{tabular}
}
%\vspace{-0.35cm}
\caption{\small {\tt PP} vs. \cpsm under minimum size constraint.} 
\label{table:const:stat}
\end{table*}
%%%%%%%%%%%%%%%%%%%%%%%%%%%%%%%%%%%%%%%%%%%%%%%%%%%%

\subsection{SPM under regular constraints}
Our last experiment compares \prefixCPREG against two variants of \sma:
\smap (\sma one pass) and
\smafc (\sma Full Check). 
Two datasets are considered from~\cite{Bonchi:2008}: one synthetic
dataset ({data-200k}), and one real-life dataset (Protein). 
For {data-200k}, we used two RE: 
%$\mathtt{RE}6 \equiv A^*B(B|C)D^*E$, 
\begin{itemize}
\item $\mathtt{RE}10 \equiv A^*B(B|C)D^*EF^*(G|H)I^*$, 
\item $\mathtt{RE}14 \equiv A^*(Q|BS^*(B|C))D^*E(I|S)^*(F|H)G^*R$. 
\end{itemize}

For {Protein}, we used $\mathtt{RE}2 \equiv (S|T)$ $.$ $(R|K)$
representing {\it Protein kinase C phosphorylation} 
(where $.$ represents any symbol).  
Fig.~\ref{fig:reg} reports CPU-times comparison. 
On the synthetic dataset, our approach is very effective. 
For $\mathtt{RE}14$, our
method is more than an order of magnitude faster than \sma. 
On Protein, the gap between the $3$ methods
shrinks, but our method remains effective. For the particular
case of $\mathtt{RE}2$, the {\tt Regular} constraint can  be substituted
by restricting the domain of the first and third variables to
$\{S,T\}$ and $\{R,K\}$ respectively (denoted as \prefixCPREL), thus
improving performances. 

\begin{figure*}[t]
{\footnotesize
\begin{tabular}{ccc}
data-200k (RE10) & data-200k (RE14) & Protein (RE2) \\  
\includegraphics[width=4cm, height=3cm]{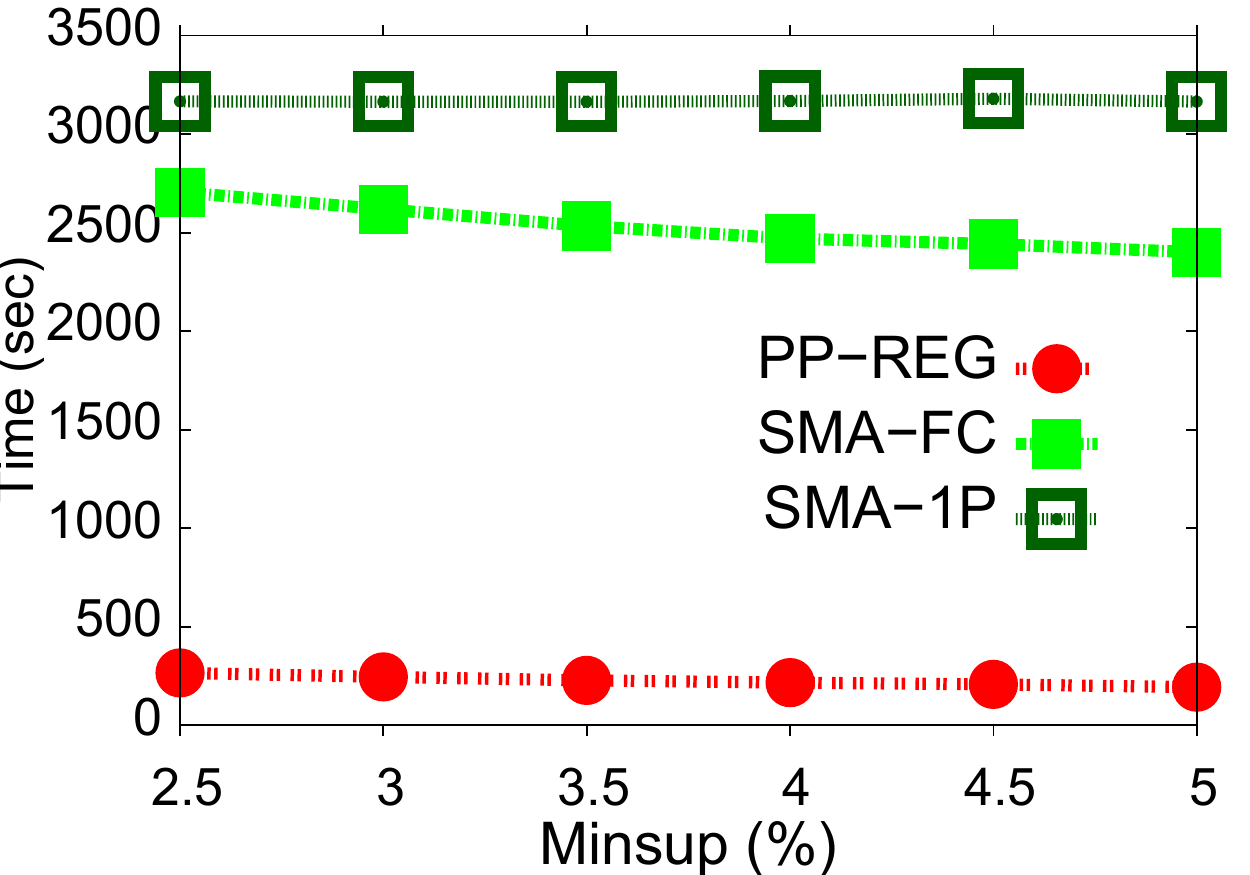}
&   
\includegraphics[width=4cm, height=3cm]{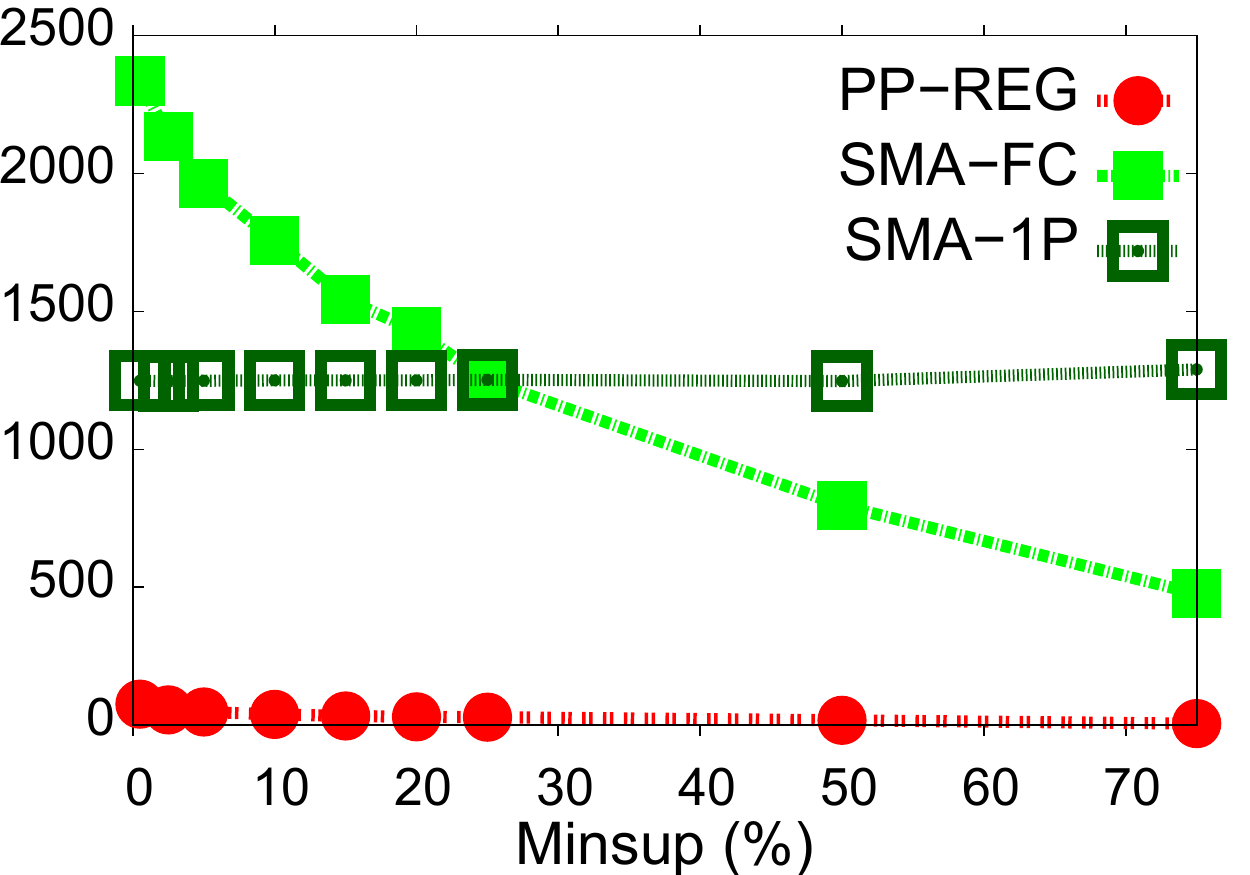}
&
\includegraphics[width=4cm, height=3cm]{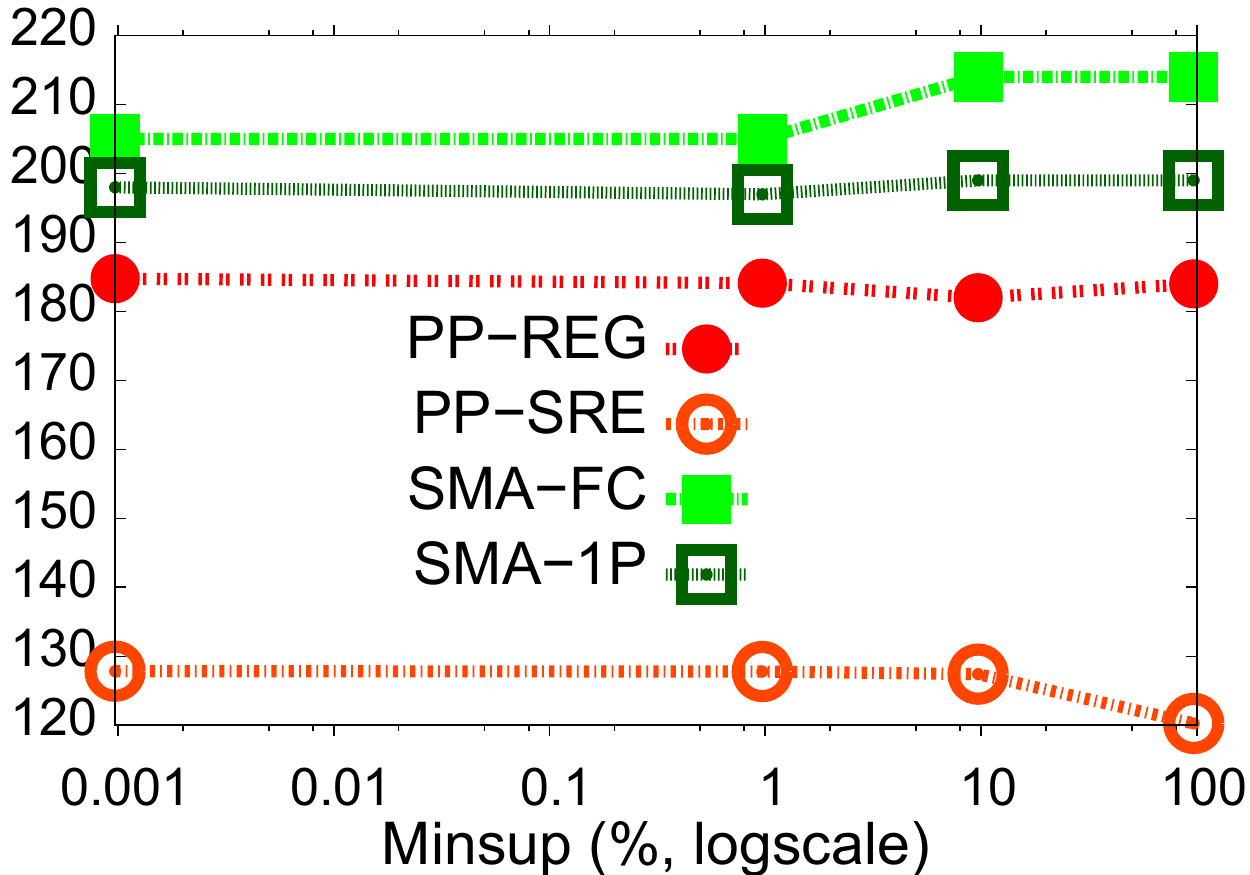}
%&
%\includegraphics[width=3cm, height=2.6cm]{figs/Freq-FIFA-GAP-col.pdf}
	\end{tabular}}
%\vspace*{-.35cm} 
%\caption{\small \label{fig:reg} Comparing \prefixCPREG with \sma at different minimum
%support thresholds, datasets and regular expressions.}
\vspace*{-.35cm} 
\caption{\small \label{fig:reg} Comparing \prefixCP with \sma for SPM under RE constraint.}
%\prefixCPREG with \sma for different regular expressions, and \prefixCPGAP with cSpade when $gap_{(0,2)}$.}
\end{figure*}

\section{Conclusion}
\label{section:conclusion}
We have proposed 
the global constraint \prefixCP for sequential pattern mining. 
\prefixCP uses a concise encoding and provides an efficient filtering 
based on specific properties of the projected databases, and
anti-monotonicity of the frequency constraint. 
When this global constraint is integrated into a CP solver, it 
enables to handle several constraints simultaneously. Some of them
like size, item membership and regular 
expression are considered in this paper. Another point of strength, is
that, contrary to existing  
CP approaches for SPM, our global constraint does not require any
reified constraints nor any extra  
variables to encode the subsequence relation. Finally, although
\prefixCP is well suited for constraints on  
sequences, it would require to be adapted to handle constraints on
subsequence relations like gap. 

Experiments performed on several real-life datasets show that our
approach clearly outperforms existing CP approaches and competes well
with ad hoc methods on large datasets and achieves scalability while
it is a major issue of CP approaches. 
As future work, we intend to handle constraints on set of sequential
patterns such as closedness, relevant subgroup and skypattern
constraints. 

%for mining frequent sequential patterns as well as for richer patterns as top-$k$ or regular sequential patterns.
%a new global constraint for mining sequential patterns in a sequence database.
%This constraint allows to capture both sub-sequence relation and frequency constraint.
% To the best of our knowledge, it is the first global constraint that exploits the prefix projection to encode the subsequence relation directly on the data.
%
%Therefore, we have exploited the prefix projection technique, used in the state-of-the art {\tt PrefixSpan} algorithm, to make an efficient filtering of domain variable. When this constraint is integrated in the machinery of constraint programming, it enables to handle several constraints simultaneously. Some of their like frequency, size and item and regular expression are considered in this paper, other constraints like gap will be studied in further works. Another point of strength, is that our idea is easily applicable to other CP solvers. 
%
%Experiments performed on different datasets shows that our global constraint is competitive with existing approaches, its combination with other constraints, shows the interest of our CP approach. 
%For future work, 
%We intend to handle constraints on set of sequential patterns 
%such as closedness, relevant subgroup and skypattern constraints. We intend also to propose a more efficient method to extract \topk patterns. 

\newpage
\bibliographystyle{splncs03}
\bibliography{seqcp15}

\end{document}